\documentclass[10pt,twocolumn,letterpaper]{article}

\usepackage[pagenumbers]{cvpr} 

\usepackage{graphicx}
\usepackage{amsmath}
\usepackage{amssymb}
\usepackage{booktabs}

\usepackage[pagebackref,breaklinks,colorlinks]{hyperref}

\usepackage[capitalize]{cleveref}
\crefname{section}{Sec.}{Secs.}
\Crefname{section}{Section}{Sections}
\Crefname{table}{Table}{Tables}
\crefname{table}{Tab.}{Tabs.}

\newcommand{\methodname}{HR-NeuS}

\newcommand{\transpose}{T}

\newcommand{\sdf}{f}
\newcommand{\mlp}{\text{MLP}}

\newcommand{\ray}{\mathbf{r}}
\newcommand{\rayorigin}{\mathbf{o}}
\newcommand{\raydir}{\mathbf{d}}

\newcommand{\position}{\mathbf{x}}

\newcommand{\tval}{t}

\newcommand{\normalvec}{\mathbf{n}}

\newcommand{\opacity}{\alpha}

\newcommand{\numfrequencies}{N}
\newcommand{\posenc}{\gamma_{pos}}
\newcommand{\hashenc}{\gamma_{hash}}

\newcommand{\mpage}[2]
{
\begin{minipage}{#1\linewidth}\centering
#2
\end{minipage}
}

\begin{document}

\title{\methodname: Recovering High-Frequency Surface Geometry via Neural Implicit Surfaces}

\author{Erich Liang\\
UC Berkeley\\
Berkeley, California\\
{\tt\small eliang@berkeley.edu}
\and
Kenan Deng\\
Amazon\\
Palo Alto, California\\
{\tt\small kenanden@amazon.com}
\and 
Xi Zhang\\
Amazon\\
Palo Alto, California\\
{\tt\small xizhn@amazon.com}
\and 
Chun-Kai Wang\\
Amazon\\
Palo Alto, California\\
{\tt\small ckwang@amazon.com}
}
\maketitle

\begin{abstract}
Recent advances in neural implicit surfaces for multi-view 3D reconstruction primarily focus on improving large-scale surface reconstruction accuracy, but often produce over-smoothed geometries that lack fine surface details. To address this, we present High-Resolution NeuS (\methodname), a novel neural implicit surface reconstruction method that recovers high-frequency surface geometry while maintaining large-scale reconstruction accuracy. We achieve this by utilizing (i) multi-resolution hash grid encoding rather than positional encoding at high frequencies, which boosts our model's expressiveness of local geometry details; (ii) a coarse-to-fine algorithmic framework that selectively applies surface regularization to coarse geometry without smoothing away fine details; (iii) a coarse-to-fine grid annealing strategy to train the network. We demonstrate through experiments on DTU and BlendedMVS datasets that our approach produces 3D geometries that are qualitatively more detailed and quantitatively of similar accuracy compared to previous approaches.
\end{abstract}

\section{Introduction}
\label{sec:intro}

Multi-view 3D surface reconstruction is a fundamental computer vision task with wide applications, such as computer graphics and virtual reality. While traditional Multi-View Stereo (MVS) approaches have been widely used, they consist of multiple stages and often fail at recovering complex scenes containing irregular or thin structures. Recently, neural surface reconstruction methods like NeuS \cite{neus} have become popular because they utilize coordinate-based multi-layer perceptrons (MLPs) to implicitly represent a 3D scene's signed distance function (SDF) and radiance field. Thanks to the MLPs, the learned geometries can be complex in structure, but are prone to local minimal solutions and often lack finer surface details compared to traditional MVS approaches. To address these issues, one recent work supervises learning by warping local patches across viewpoints and enforcing patch similarity \cite{francois2022neuralwarp}, but this implicitly assumes that local surface geometry is planar, leading to even less surface geometry texture. Others take a hybrid approach by using traditional MVS output as supervision \cite{Johari_2022_CVPR, geoneus}, which adds complexity and other geometric priors to the pipeline.

We propose a multi-view neural surface reconstruction approach capable of expressing high frequency surface details while maintaining accurate overall geometry, without relying on strong geometric assumptions. We represent scene geometry via SDF-based neural volumetric rendering similar to NeuS's \cite{neus} formulation. We determine through theoretical analysis and empirical results that using multi-resolution hash encoding in place of positional encoding improves overall expression of high frequency details. To maintain the fidelity of the overall geometry, we optimize our model using a coarse-to-fine annealing strategy, and apply regularization at the coarsest geometry levels to promote piece-wise smooth surface formation without impacting fine surface details. Experimental evaluation on the DTU \cite{dtu} and BlendedMVS \cite{blendedmvs} datasets show that our algorithm achieves similar levels of accuracy compared to state of the art, while having more fine-detailed surface structure.

\section{Related Work}
\label{sec:relatedwork}
\noindent \textbf{Multi-view object reconstruction.} Traditional Multi-View Stereo (MVS) methods estimate depth information by maintaining a point-based \cite{barnes2009patchmatch, furukawa2009accurate, galliani2016gipuma, schonberger2016pixelwise, campbell2008using, tola2012efficient} or a surface-based \cite{de1999poxels, broadhurst2001probabilistic, seitz1999photorealistic} matching across different views. In \cite{leroy2018shape} photoconsistency among views are imposed. Such consistencies between views are enforced by post-processing steps of depth fusion \cite{curless1996volumetric, merrell2007real, donne2019learning, riegler2017octnetfusion} followed by mesh surface reconstruction processes such as Poisson surface reconstruction \cite{kazhdan2006poisson} and ball-pivoting \cite{bernardini1999ball}. Recent approaches train end-to-end systems either with supervision \cite{yao2018mvsnet, huang2018deepmvs}, or without supervision but under multi-view consistency constraints \cite{dai2019mvs2, khot2019learning, huang2021m3vsnet, xu2021self}. Many of these end-to-end systems suffer from low resolution of reconstructed objects due to memory constraints.
\begin{figure*}[t!]
     \centering
     \includegraphics[width=1\textwidth]{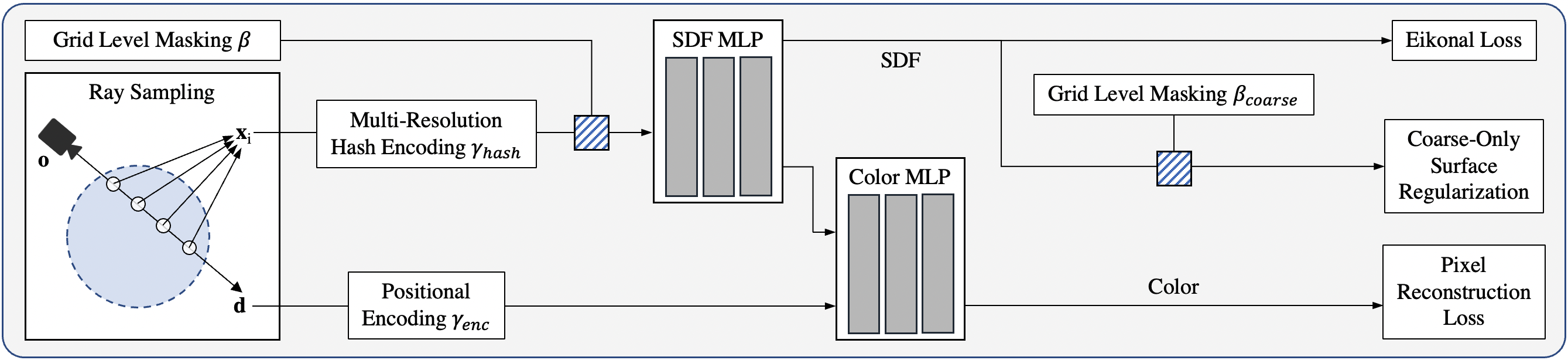}
   \caption{Overview of \methodname. Previous neural implicit surface methods fail to capture high-frequency surface geometry. To address this, we utilize multi-resolution hash encoding to help increase the local expressiveness of the surface learned by our model. To ensure that overall geometry is still correct, we introduce a novel coarse-to-fine framework that consists of a special regularization technique and coarse-to-fine annealing. These encourage piece-wise smooth surface formation on a coarse level without impacting the fine details learned on the surface.}
\label{fig:overview}
\end{figure*}

\noindent \textbf{Neural implicit surfaces.} Discretized representations such as mesh, point cloud, and voxels usually suffer from discretization artifacts. To overcome such drawbacks, implicit functions have drawn attention in last two decades in modeling 3D geometry \cite{niessner2013real, YOO2011934, 10.1145/2732197, WANG2011463, 10.1145/2601097.2601163} due to expressive representation power and low memory footprint. Implicit representation also demonstrates its capability on prediction of appearance \cite{liu2020neural, liu2020dist, dvr, oechsle2019texture, oechsle2020learning}. Among these works, IDR \cite{idr} fits a zero level-set representation using differentiable render to represent object surface and determines light radiance directly on the surface. A similar idea was adopted by DVR \cite{dvr}. However, these approaches are hard to optimize and require segmentation mask supervision, which our method does not. NeRF \cite{nerf} inspired a direction of research that combines neural implicit representation and volume rendering, which has been optimized by some works to produce surface geometry output. UNISURF \cite{unisurf} learns an occupancy grid representation for an object by training a differentiable render that combines the merit of both surface rendering and volume rendering. Rather than representing an object by occupancy values, NeuS \cite{neus} uses an SDF function in representation which enables the approach to capture more object details in reconstructed results. Similarly, an SDF transformed by Laplace's cumulative distribution function is used in VolSDF \cite{volsdf}. Recent works \cite{francois2022neuralwarp, geoneus} have added additional regularization terms to improve reconstruction accuracy. However, these works do not recover high-frequency surface details, which our method does.

\noindent \textbf{Grid-based encoding.} Partitioning space with grids that represent information at arbitrary spatial locations via interpolation has proven to be effective for neural implicit encoding. Based on the observation that different objects can share similar part geometries, \cite{jiang2020local} reconstructs a target in a large scale setup by training an autoencoder to learn implicit latent encoding of local grids sampled from artificial 3D models. The trained decoder part then extracts targeted geometry from sparse point samples.  \cite{convoccnet} uses a convolutional autoencoder to learn latent features, which are decoded to predict object occupancy in occupancy grids. The convolution is translational equivariant and encodes geometry information in a multiscale and hierarchical manner in different network layers, thus enabling better structure reasoning. Rendering complex shapes with fine details usually requires more granular grid representations, which is computationally expensive and hard to scale up. To solve this problem, level of detail (LOD) framework combined with implicit representation is adopted by many recent works. In \cite{takikawa2021neural}, given a learned geometry prior, Takikawa \textit{et al.} proposes a multiscale octree based representation using voxels to represent implicit surfaces \cite{barron2021mip}. \cite{Muller_2022} proposes multi-resolution hash encoding, which is more efficient in memory and does not rely on a geometry prior.

\section{Method}
\label{sec:method}

Given posed 2D images of an object, our algorithm produces a 3D reconstruction of the object's surface that captures high-frequency surface geometry details while maintaining accurate overall geometry. Inspired by \cite{neus}, we represent object geometry with an implicit SDF MLP and optimize it via neural volumetric rendering (\textsection \ref{subsec:nvr}). Within this setup, we hypothesize that positional encoding limits the local expressiveness of learned geometry, so we use multi-resolution hash encoding \cite{Muller_2022} instead (\textsection \ref{subsec:multires}). However, this increases the likelihood of local minima solutions, so we introduce a novel coarse-to-fine algorithmic framework which consists of a special surface regularization technique (\textsection \ref{subsec:novelloss}) and a coarse-to-fine annealing strategy (\textsection \ref{subsec:c2f}), which together encourage the formation of large-scale piece-wise smooth geometry without removing high-frequency surface details. Training details of our algorithm are presented in \textsection \ref{subsec:training}. An overview of our method is presented in Figure \ref{fig:overview}.

\subsection{SDF-Based Neural Volumetric Rendering}
\label{subsec:nvr}

Given a 3D object with a watertight surface, its SDF is a function $\sdf: \mathbb{R}^3 \rightarrow \mathbb{R}$ that takes in any point $\position \in \mathbb{R}^3$ and outputs the signed distance between $\position$ and the closest point on the surface. Importantly, the object surface $\mathcal{S}$ is the zero-level set of the SDF:
\begin{align}
    \mathcal{S} = \{\position \in \mathbb{R}^3 \ | \ \sdf(\position) = 0\}.
    \label{eq:sdf}
\end{align}
Because the SDF precisely describes an object's geometry, recent works utilize SDF-based neural volumetric rendering \cite{neus, volsdf} to extract cleaner surfaces from input sets of 2D posed images. The main idea is to use a system of MLPs to implicitly describe a 3D scene's SDF and view-dependent color properties. These MLPs' outputs are then fused via differentiable volumetric rendering to render 2D images of the scene, which should look similar to the input images. To optimize these MLPs, backpropagation is applied to the $\ell_1$ color loss between reconstructed pixels and original pixels.

To synthetically render the color of a pixel, we sample points
points $\position_i = \ray(\tval_i) = \rayorigin + \tval_i \raydir$ along the ray that starts at the camera's center of projection $\rayorigin$ and passes through the pixel with unit direction $\raydir$. We only consider points within the unit sphere, because we assume that the scene has been rescaled to fit within this region like \cite{neus}. We approximate the rendered pixel color as a weighted sum of the colors of points sampled along the ray, where the weight of each point depends on its SDF value; this is similar to density-based volumetric rendering described in \cite{nerf}, except we replace densities with SDF-dependent opacity values $\opacity_i$, and we refer to \cite{neus} for the exact conversion from SDF values to opacity values.

Before querying the colors and SDFs of points $\position_i$ along the ray, we must apply encoding to the coordinates of $\position_i$; this is done to help MLPs learn higher frequency functions \cite{tancik2020fourfeat}, such as SDF and color. Many works \cite{nerf, tancik2020fourfeat, neus} utilize positional encoding:
\small
\begin{equation}
    \gamma(\position)\!=\!\Big[ \sin(\position), \cos(\position), \ldots, \sin\!\big(2^{\numfrequencies-1} \position\big), \cos\!\big(2^{\numfrequencies-1} \position\big) \Big]^\transpose
    \label{eq:posenc}
\end{equation}
\normalsize
However, we utilize multi-resolution hash encoding for 3D spatial coordinates (see \ref{subsec:multires}). We still use positional encoding for viewing direction $\raydir$, so we denote these encoding schemes as $\posenc$ and $\hashenc$. The SDFs, opacities, and colors queried from the MLPs can be written as
\begin{align}
    \sdf_i &= \mlp(\hashenc(\position_i); \Theta_{sdf})
    \\
    \opacity_i &= g(\sdf_i)
    \\
    c_i &= \mlp(\hashenc(\position_i), \normalvec_i, \posenc(\raydir); \Theta_c)
\end{align}
Here, $\normalvec_i$ is the direction of the surface normal which is the gradient of the SDF, $g$ is the conversion from SDF to opacity values from \cite{neus}, and $\Theta_{sdf}$ and $\Theta_c$ are the learned weights of each MLP. These values can be used in classical volume rendering equations to produce the final rendered pixel color
\begin{equation}
    \hat{C} = \sum_{i = 1}^n T_i \opacity_i c_i
\end{equation}
where $T_i$ is the discretized accumulated transmittance defined as $T_i = \prod_{j = 1}^{i - 1}(1 - \opacity_j)$.

\subsection{Benefits of Multi-Resolution Hash Encoding}
\label{subsec:multires}

In this section, we present analysis and empirical results to explain why choose to apply multi-resolution hash encoding \cite{Muller_2022} instead of positional encoding to 3D spatial coordinates in our algorithm. Before we do so, we provide a description of how multi-resolution hash encoding works within the context of our neural volumetric renderer.

First, we define the domain for the encoding as an axis-aligned cube $[-1, 1]^3 \in \mathbb{R}$; this contains the unit sphere which the 3D scene resides in. We cover this cube with a total of $L$ levels of lattice grids of different resolutions. The first level ($l = 1$) has the coarsest resolution of $N_{\min}$ points along one of the cube's sides, and each subsequent level $l$ has a resolution of $N_l = N_{\min} \cdot 2^{l - 1}$. In each level, each grid point is mapped to a ``feature fragment'' of length $d$. To save on memory, each level has a maximum number of distinct feature fragments it can have; hence, for higher level grids, some of its grid points may collide and map to the same feature fragment.

\begin{figure}[t]
  \centering
   \includegraphics[width=0.8\linewidth]{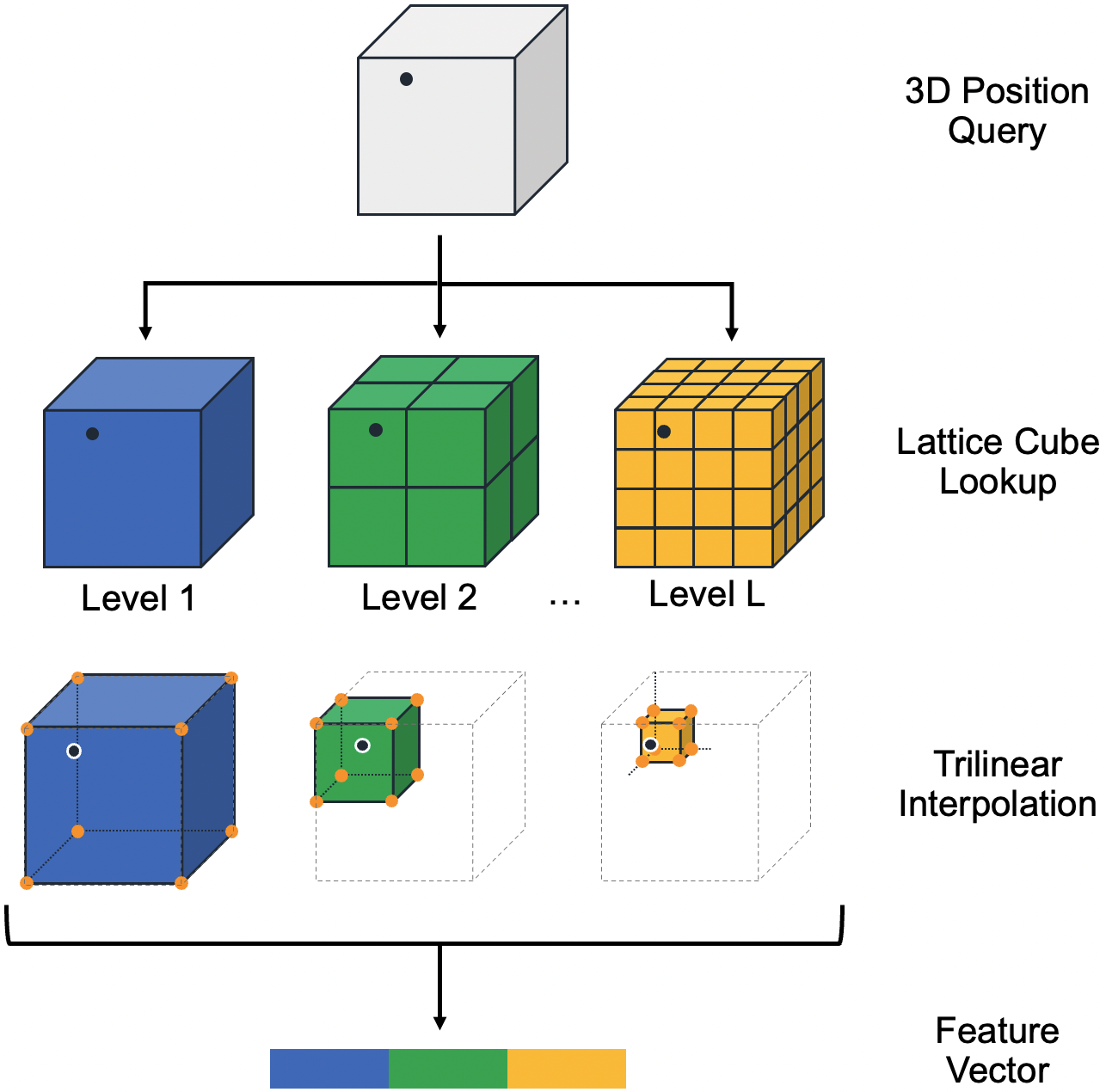}
   \caption{An illustration of how multi-resoution hash encoding converts an input 3D position into a feature vector. Each grid level's feature fragments, which are stored on individual lattice grid points, are learnable and independent of each other, allowing the encoding to adapt its encoding to help an MLP best learn the underlying signal downstream.}
   \label{fig:multireshash}
\end{figure}

We now encode any input $\position \in [-1, 1]^3$ as follows: for each grid level $l$, we locate the lattice cube that contains $\position$. For each lattice cube, we query the feature fragments stored in its 8 corners, and apply trilinear interpolation over the feature fragments to form a final feature fragment $F_l \in \mathbb{R}^d$ that describes $\position$. We then concatenate each grid level's feature fragments together to form the final feature vector: $\hashenc(\position) = \overline{F_1 F_2 \dots F_L} \in \mathbb{R}^{L \times d}$. See Figure \ref{fig:multireshash} for an illustration of this process.

Positional encoding's frequency bands and multi-resolution hash encoding's grid levels play a similar role in terms of how they help a downstream MLP learn. Consider a coordinate-based MLP that takes in as input $\position \in \mathbb{R}^3$ and needs to approximate a ground-truth signal $h(\position)$. Coordinate-based MLPs without the aid of any encoding scheme have a spectral bias towards low frequencies \cite{tancik2020fourfeat}, meaning that they converge much faster when learning a $h(\position)$ varies slowly as $\position$ varies, i.e. $h(\position)$ is of low frequency, but struggle to converge when learning higher frequency $h(\position)$. NTK analysis shows that the higher frequency bands in positional encoding help MLPs converge faster when learning high frequency signal \cite{tancik2020fourfeat}. Similarly, higher resolution grid levels in multi-resolution hash encoding can help MLPs better learn high frequency signal as well. High resolution grid levels have smaller lattice grids which can have arbitrary feature fragments attached to their corners. Hence, as $\position$ varies over $\mathbb{R}^3$ space, higher grid levels have the ability to generate higher frequency feature fragments. But then, the map from the high frequency space of $\hashenc(\position)$ to the signal $h(\position)$ may no longer be a high-frequency function anymore, which could improve an MLP's effectiveness at learning $h(\position)$.

However, we argue that MLPs with multi-resolution hash encoding are still overall more capable of learning high frequency signals compared to MLPs with positional encoding. Consider any two distinct points $\position_1$ and $\position_2$ in 3D space such that $h(\position_1) \neq h(\position_2)$. We claim that positional encoding's high frequency bands are more likely to map $\position_1$ and $\position_2$ to similar values in encoding space compared to multi-resolution hash encoding's high resolution grid levels. To see why, note that the frequency bands in positional encoding are always forced to output a value in the range $[-1, 1]$, whereas the feature fragments generated by multi-resolution hash encoding can span all of $\mathbb{R}^d$. In addition, positional encoding is periodic, so its frequency bands are each guaranteed to map some 3D points to the same encoding value; this occurs even more often for higher frequency bands. In contrast, multi-resolution hash encoding utilizes trainable feature fragments that can adapt to best fit the underlying signal the MLP is trying to learn. Hence, due to positional encoding's inherently smaller range, periodic nature, and inability to adapt its encoding to the underlying signal, it is more likely for the encoding produced by high frequency bands of $\position_1$ and $\position_2$ to collide compared to high resolution grid levels. But since the high frequency bands in positional encoding are responsible for helping the MLP learn high frequency signals, this ultimately results in the issue of ``global entanglement'' of high frequency details in the MLP's prediction of the ground truth signal, leading to worse expressiveness of high frequency signals.

We test our claim empirically by comparing the surface reconstruction quality of SDF-based neural volumetric rendering methods that either apply purely positional encoding or purely multi-resolution hash encoding to spatial 3D coordinates. We pick the parameters of both encoders to create as fair of a comparison as possible. Specifically, notice that one can think of positional encoding as a special case of multi-resolution hash encoding, where each frequency band can be approximated by a grid layer with feature fragment length of $6$ (two per each $(x, y, z)$ coordinate), fixed feature fragments that take on values of either 0, 1, or -1, and utilizes ``sinusoidal interpolation'' within the lattice grid cells. To compute the resolution of the grid level that would correspond to the $n$th frequency band $\big(\sin(2^{n-1} \position), \cos(2^{n-1} \position)\big)$, note that the $n$th frequency band has a period of $\frac{2 \pi}{2^{n - 1}}$ that occurs $\frac{2^{n - 1}}{\pi}$ times within the interval $[-1, 1]$. Each frequency band period can be approximated by doing linear interpolation between 4 points with the values of 0, 1, 0, and -1; hence, we can utilize a layer from multi-resolution hash encoding with resolution $4 \cdot \frac{2^{n - 1}}{\pi} = \frac{2^{n + 1}}{\pi}$ to achieve similar spatial resolution as the frequency band from the positional encoder.

\begin{figure}[t]
  \centering
   \includegraphics[width=0.48\linewidth]{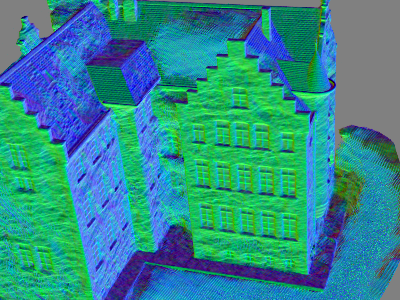}
   \includegraphics[width=0.48\linewidth]{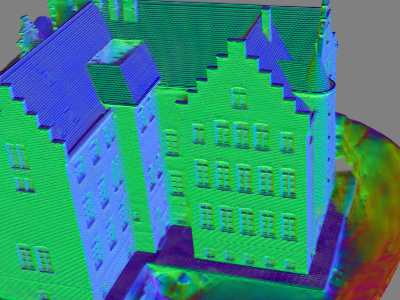}
   \caption{The normal maps generated from surface reconstruction utilizing either positional encoding (left) or multi-resolution hash encoding (right). Parameters of each encoder were picked to give each one similar spatial resolution; all other hyperparameters and network structures were kept the same.}
   \label{fig:posvgrid}
\end{figure}

Using these calculations, we run two trials of surface reconstruction of scene 24 from the DTU dataset; first, we utilize a positional encoder with $N = 11$ frequency bands, and second, we utilize a multi-resolution hash encoder with $l = 11$ levels, where the first level has grid resolution of $N_{\min} = \frac{4}{\pi}$ and feature fragment length of $d = 4$ per level; ideally, we set $d = 6$, but due to constraints from \cite{tiny-cuda-nn}, we are only able to choose perfect powers of 2 for $d$. As expected, the surface generated via positional encoding exhibits worse high frequency details compared to the surface generated via multi-resolution hash encoding (see Figure \ref{fig:posvgrid} for each trial's normal map). In particular, positional encoding generates a bumpy roof with streak patterns, whereas multi-resolution hash encoding generates an overall smooth roof in which individual roof tiles can be clearly seen.

\subsection{Coarse-Only Surface Regularization}
\label{subsec:novelloss}

\begin{figure}[t]
  \centering
    \includegraphics[width=0.48\linewidth]{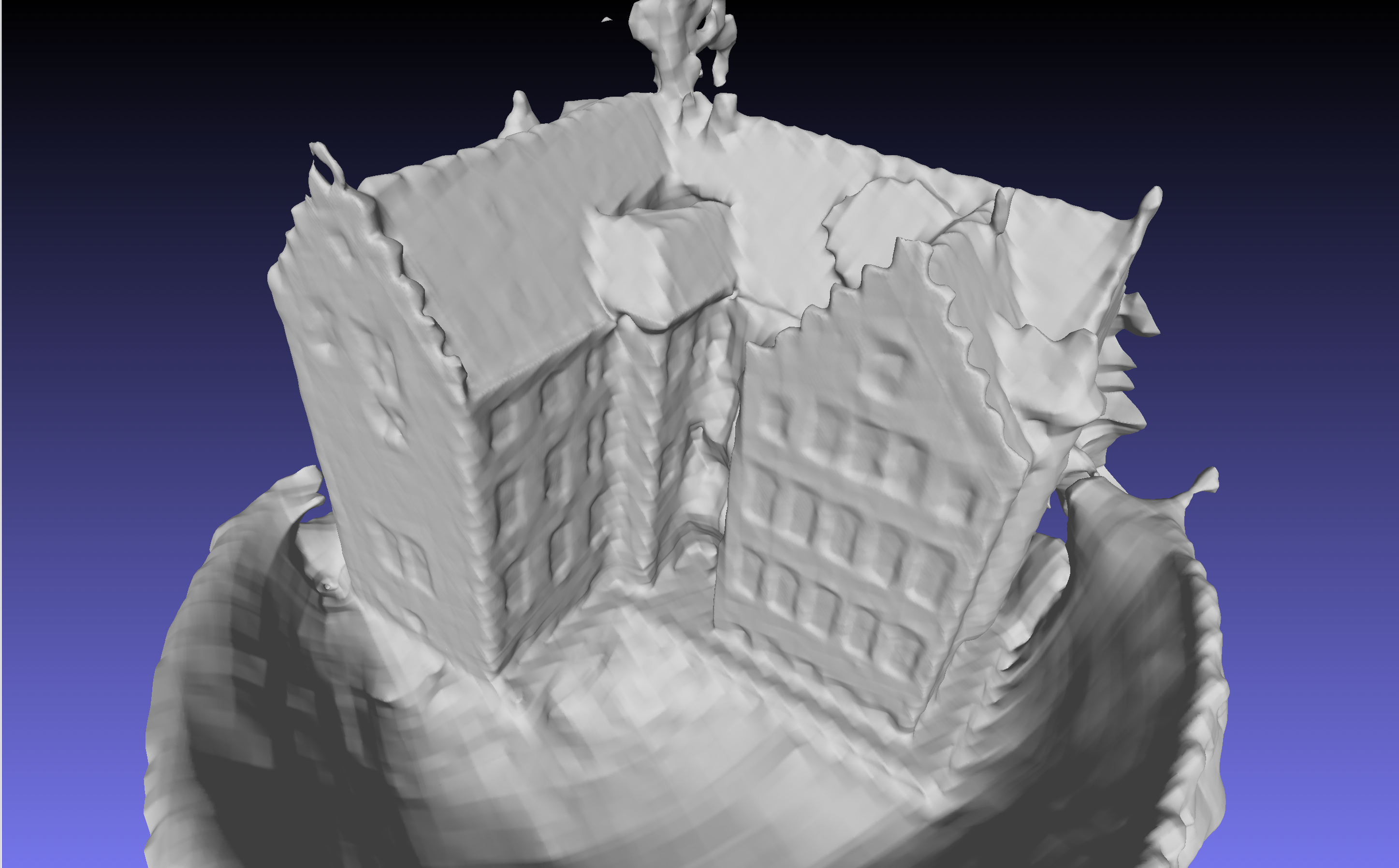}
   \includegraphics[width=0.48\linewidth]{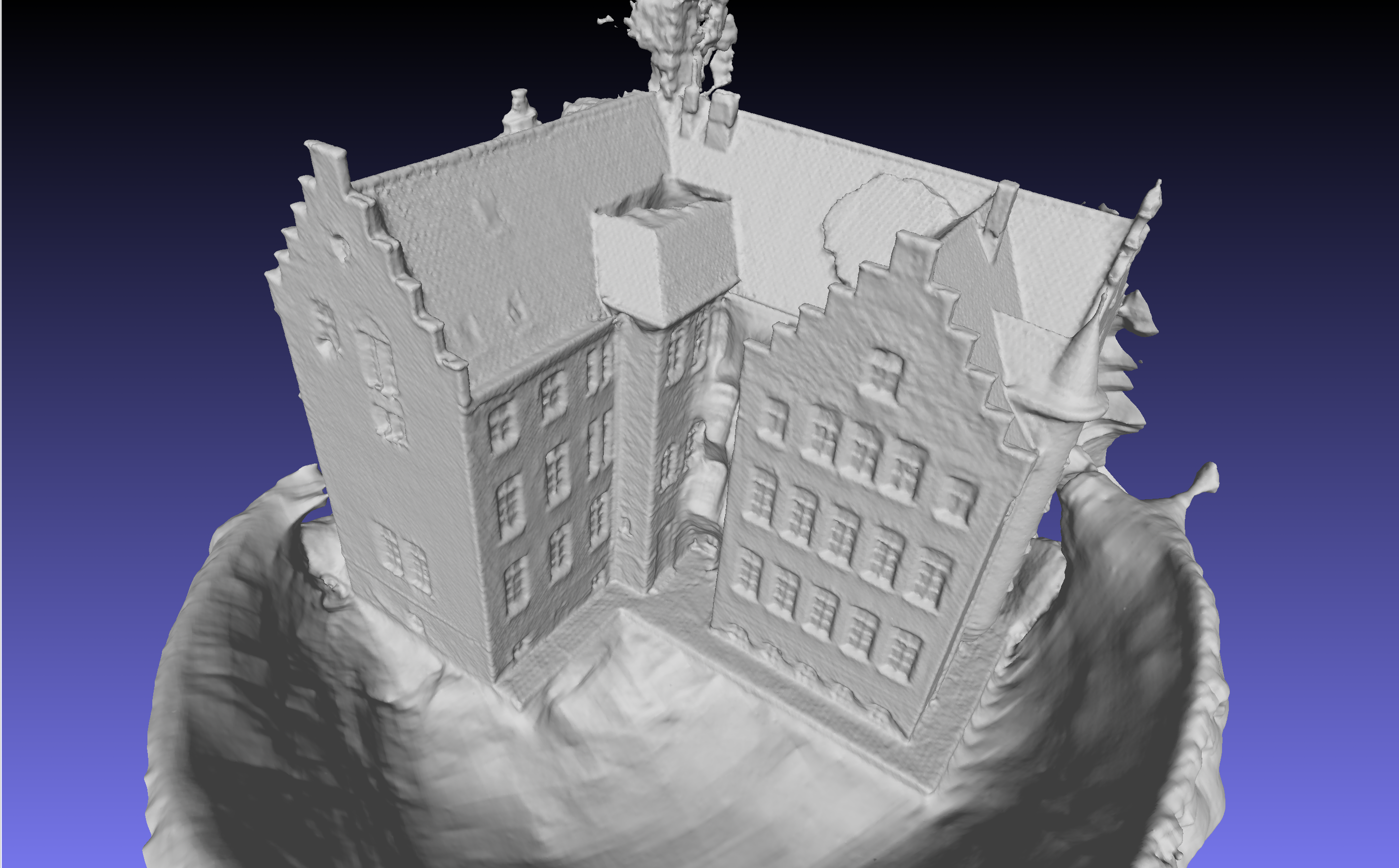}
   \caption{An example of the local minima surface geometry that can form when training using multi-resolution encoding directly. The left mesh is extracted using $\hashenc(\position, \beta_{\text{coarse}} = 4)$, which has higher resolution grid levels zeroed out. The right mesh is generated using all levels of the multi-resolution hash encoding.}
   \label{fig:bump}
\end{figure}

Although utilizing multi-resolution hash encoding does improve our model's ability to recover high-frequency features, it simultaneously increases the total number of trainable parameters in our model. This makes the optimization problem more underconstrained and increases the likelihood our model falls into local minima, especially when we use a large number of grid levels simultaneously. As a result, artifacts can appear in the learned surface, such as the large bump that appears in Figure \ref{fig:bump}.

One approach for avoiding such local minima is to add regularization terms that encourage more desirable types of surface to form. Typically, these strategies encourage the formation of smooth or piece-wise smooth surfaces. However, applying these types of regularization is directly at odds with the goal of learning high-frequency details in the 3D reconstruction of the surface, as the regularization will actively smooth out any fine details recovered by our model.

To solve this dilemma, we propose a novel regularization technique that does not directly act on the entire learned surface, but rather on a coarse version of it generated only using the lower resolution grid levels in our multi-resolution hash encoder. Specifically, we expand our definition of $\hashenc$ to take in an additional parameter $\beta$ \cite{nerfies} that helps softly zero out the values of the feature fragments generated by higher resolution grid levels:
\begin{align}
    \hashenc(\position, \beta) = \overline{F_1' F_2' \dots F_L'}\ \text{where}
    \\
    F_i' = w_i(\beta) F_i\ \text{and}
    \\
    w_i(\beta) = \frac{1 - \cos(\pi \cdot \text{clamp}(\beta - i + 1, 0, 1))}{2}
    \label{eq:multires}
\end{align}
Intuitively, grid layer $F_i$ will be fully activated if $i \leq \beta$, grid layer $F_{\lceil \beta \rceil}$ will be partially activated, and all other higher grid layers will be zeroed out. Using this notation, our surface regularization technique to apply the surface regularization term to the surface that is generated by the SDF MLP when using $\hashenc(\position, \beta_{\text{coarse}})$ as the encoder, where $1 \leq \beta_{\text{coarse}} < L$ is a hyperparameter; denote this surface as $\mathcal{S}_{coarse}$. The intuition behind this choice is that when utilizing multi-resolution hash encoding, the lower-indexed grid levels contain the most information about the coarse shape of the surface, while the higher-indexed grid levels primarily contain information about the high-frequency details on the surface (see Figure \ref{fig:bump}). As a result, applying surface regularization to $\mathcal{S}_{coarse}$ will safely and effectively apply any smoothing surface regularization we choose without impacting the high-frequency surface detailed learned in higher resolution grid levels.

\begin{figure}
\begin{minipage}{0.4\linewidth}
    \centering
    \includegraphics[width=\linewidth]{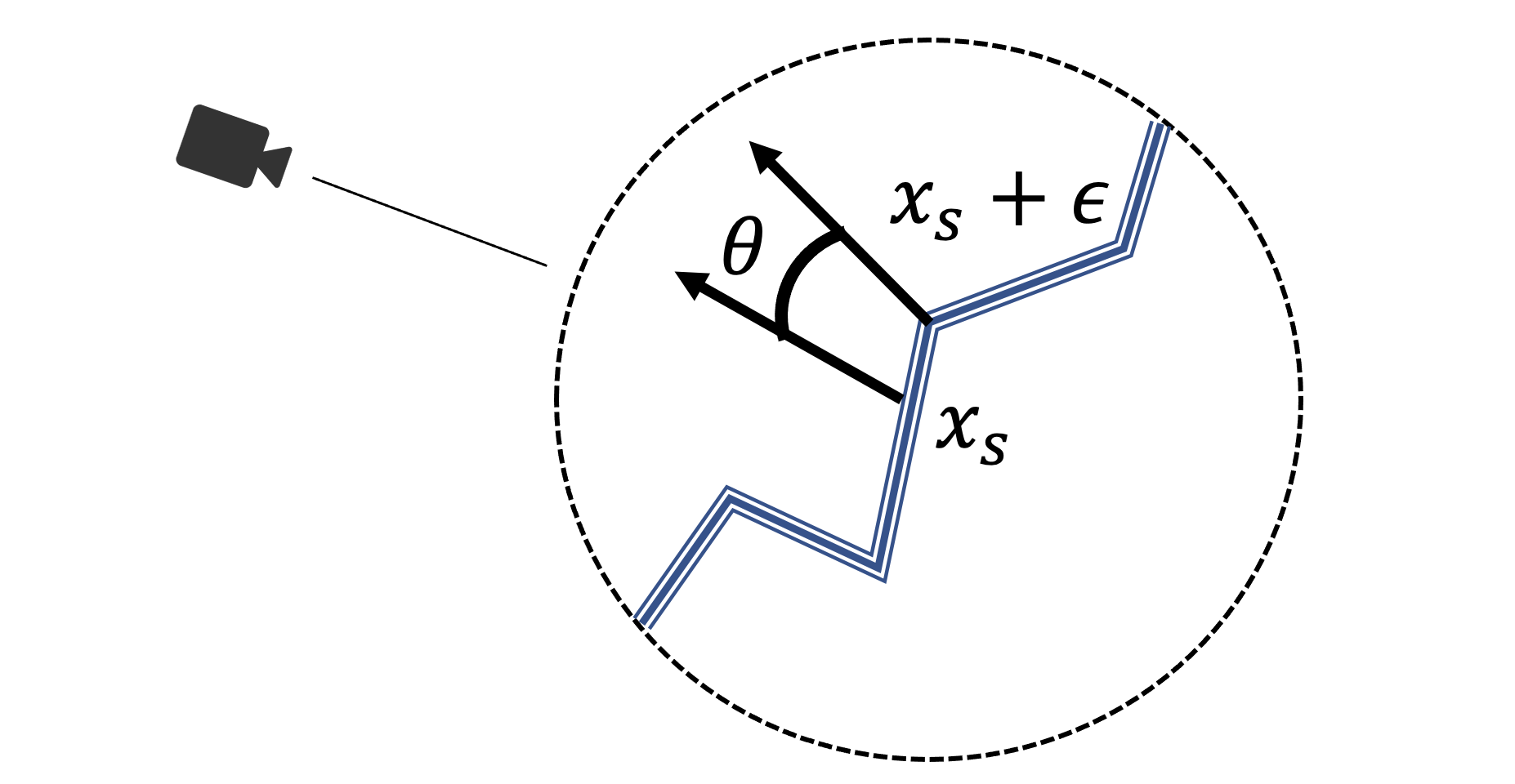}
\end{minipage}
\begin{minipage}{0.4\linewidth}
 \begin{equation*}
      w = \begin{cases} 
          1 & \theta < \theta_{\min} \\
          \frac{\theta_{\max} - \theta}{\theta_{\max} - \theta_{\min}} & \theta \in [\theta_{\min}, \theta_{\max}] \\
          0 & \theta > \theta_{\max}
       \end{cases}
 \end{equation*}
\end{minipage}
\caption{Our weighted normal regularization formulation.}
\label{fig:normal_loss}
\end{figure}

For our work, we choose to utilize a simple surface regularization term that encourages piece-wise smooth surface formation for the surface generated by the coarse grid levels. Specifically, we define
\begin{equation}
    \mathcal{L}_{reg, \beta_{\text{coarse}}} = \sum_{\position_s \in \mathcal{S}} w(\normalvec(\position_s), \normalvec(\position_s + \epsilon)) \cdot ||\normalvec(\position_s) - \normalvec(\position_s + \epsilon)||_2
    \label{eq:reg}
\end{equation}
where $\normalvec$ denotes the unit surface normal vector at that point, $\position_s$ is a point near the surface with SDF value less than a chosen $sdf_{threshold}$, $\position_s + \epsilon$ is $\position_s$ displaced by a small $\epsilon$ in a randomly selected direction, $w(\normalvec(\position_s), \normalvec(\position_s + \epsilon))$ is a weighting term that is linearly varies between 0 and 1 within a certain angles. See Figure~\ref{fig:normal_loss}. Intuitively, this encourages relatively flat surfaces to become flatter, while leaving sharp surface direction changes alone to help preserve the sharpness of edges between different planar parts of the learned surface. Note that although we choose to use this type of surface regularization, almost any other kind of surface regularization can be applied in the same manner to $\mathcal{S}_{coarse}$ while leaving higher frequency details alone.

\subsection{Coarse to Fine Annealing}
\label{subsec:c2f}

To further guide the training process to avoid local minima surface solutions, we employ a coarse to fine training framework inspired by \cite{nerfies}. The main idea is to first let the SDF MLP only use the lower resolution grid levels of multi-resolution hash encoding to learn a roughly correct coarse surface corresponding to the object of interest. Once the coarse level mesh has stabilized, we then slowly reintroduce higher resolution grid levels to the model. This effectively reduces the total number of parameters used during training initially and allows the model to obtain better geometry initially. Hence, when we reintroduce higher resolution grid levels to learn high frequency surface features, accurate overall geometry is already established, and we can let the model fully focus on expressing local features along the surface. Specifically, when computing $\mathcal{L}_{rec}$ during training, we do neural volumetric rendering based on $\hashenc(\position, \beta)$, where $\beta$ is a parameter that follows a schedule that starts out equal to $\beta_{\text{coarse}}$ and slowly increases to $L$.

\subsection{Training Details}
\label{subsec:training}

At the beginning of each optimization iteration, we first select a random camera and sample 512 random pixels from its image to form a minibatch of 512 rays, similar to \cite{neus}. Along each of these rays, we sample 64 points uniformly distributed between a near and far plane that roughly correspond with the intersection of the rays with the unit sphere of interest. We then sample 64 more points in total via hierarchical sampling similar to \cite{neus}. We train without mask supervision, so we sample an additional 32 points along each ray and use them to learn a separate NeRF++ \cite{nerfpp} model of the object background.

For encoding 3D positions, we utilize multi-resolution hash encoding with $L = 16$ total grid levels and coarsest resolution of $8$ points along the cube of sidelength $[-1, 1]$. We use tiny-cuda-nn \cite{tiny-cuda-nn} implementation for the multi-resolution hash encoding to speed up the training process. Each grid level is allowed up to $2^{22}$ distinct feature fragments, and each level's feature fragment is of length $d = 4$. We utilize positional encoding with 4 frequency bands for encoding view direction. The SDF MLP we use has 4 hidden layers of width 256, with a skip connection in the middle. We utilize geometric initialization similar to \cite{sal}, so our SDF network takes in a position $\position$'s original 3D coordinates alongside its encoded version. The SDF network output is of size 257, where the first value is used as the SDF value, and the next 256 values is a feature vector fed into the color network. Our color network follows the same color network found in \cite{neus}. Similar to IDR \cite{idr}, we utilize weight normalization to help stablize training.

The overall loss function we optimize is
\begin{equation}
    \mathcal{L} = \mathcal{L}_{rec} + \lambda_{eik} \mathcal{L}_{eik} + \lambda_{reg} \mathcal{L}_{reg, \beta_{\text{coarse}}}
\end{equation}

where $\mathcal{L}_{rec, \beta} = \sum_{r \in \mathcal{R}} ||\hat{C}(r) - C(r)||_1$ is the $\ell_1$ pixel reconstruction loss from every sampled minibatch of rays $\mathcal{R}$ when using $\hashenc(\position, \beta)$ as the SDF MLP's spatial encoding; $\mathcal{L}_{eik}$ is the same eikonal loss defined in \cite{eik} to make sure the learned SDF function is well defined; and $\mathcal{L}_{reg, \beta_{\text{coarse}}}$ is the surface regularization term applied to $\mathcal{S}_{coarse}$ as defined in \textsection \ref{subsec:novelloss}. We use $\lambda_{eik} = 0.1$, $\lambda_{reg} = 1 \times 10^{-4}$. For the surface regularization, we set $\beta_{\text{coarse}} = 4$ and use $\theta_{\min} = 22.5$ degrees, $\theta_{\max} = 60$ degrees, $\epsilon = 1 \times 10^{-4}$, and $sdf_{threshold} = 1 \times 10^{-3}$. We train our algorithm for a total of 160k steps; for scheduling $\beta$, we first keep it equal to $4$ for the first 32k steps, and linearly increase it so it is equal to 16 at step 128k; we keep it at 16 after that point. We train our MLPs using ADAM \cite{adam}. Like \cite{neus}, our learning rate is linearly warmed up from 0 to $5 \times 10^{-4}$ and then is changed via cosine decay scheduling to a minimal learning rate of $2.5 \times 10^{-5}$. We train each model for a total of 6 hours on a single A100.

\begin{table}
  \centering
  \begin{tabular}{c||c|c|c}
    Scan &  NeuS\cite{neus}  & NeuralWarp\cite{francois2022neuralwarp} & Ours \\
    \midrule 
    24 & 0.92 & 0.48 & 0.49 \\
    37 & 1.02 & 0.64& 0.65\\
    40 & 0.79 &  0.36 & 0.35\\
    55 & 0.36 &  0.37  & 0.34 \\
    63 & 0.99 &  0.78 & 0.86 \\
    65 & 0.56 &  0.79 & 0.57 \\
    69 & 0.53 &  0.78 & 0.56 \\
    83 & 1.39 &  1.16 & 1.33 \\
    97 & 1.14 &  1.04 & 1.18 \\
    105 & 0.75 &  0.65 & 0.70 \\
    106 & 0.54 & 0.64 & 0.53 \\
    110 & 1.11 &  0.70 & 1.04 \\
    114 & 0.34 & 0.40 & 0.37\\
    118 & 0.42 &  0.57 & 0.40 \\
    122 & 0.41 &  0.48 & 0.44 \\
    \hline
    mean & 0.75 & 0.66 & 0.65 \\
  \end{tabular}
  \caption{Results on 15 scenes in DTU~\cite{dtu} dataset. We use chamfer distance measure the reconstructed mesh. Lower is better. We use the same post-process step to clean the mesh for fair comparison.}
  \label{tab:dtu_result}
\end{table}

\begin{figure*}[!h]
 \centering
 \rotatebox[origin=C]{90}{\parbox{20mm}{\centering \small Scan 24 \\ (DTU)}} 
  \mpage{0.22}{\includegraphics[width=\linewidth]{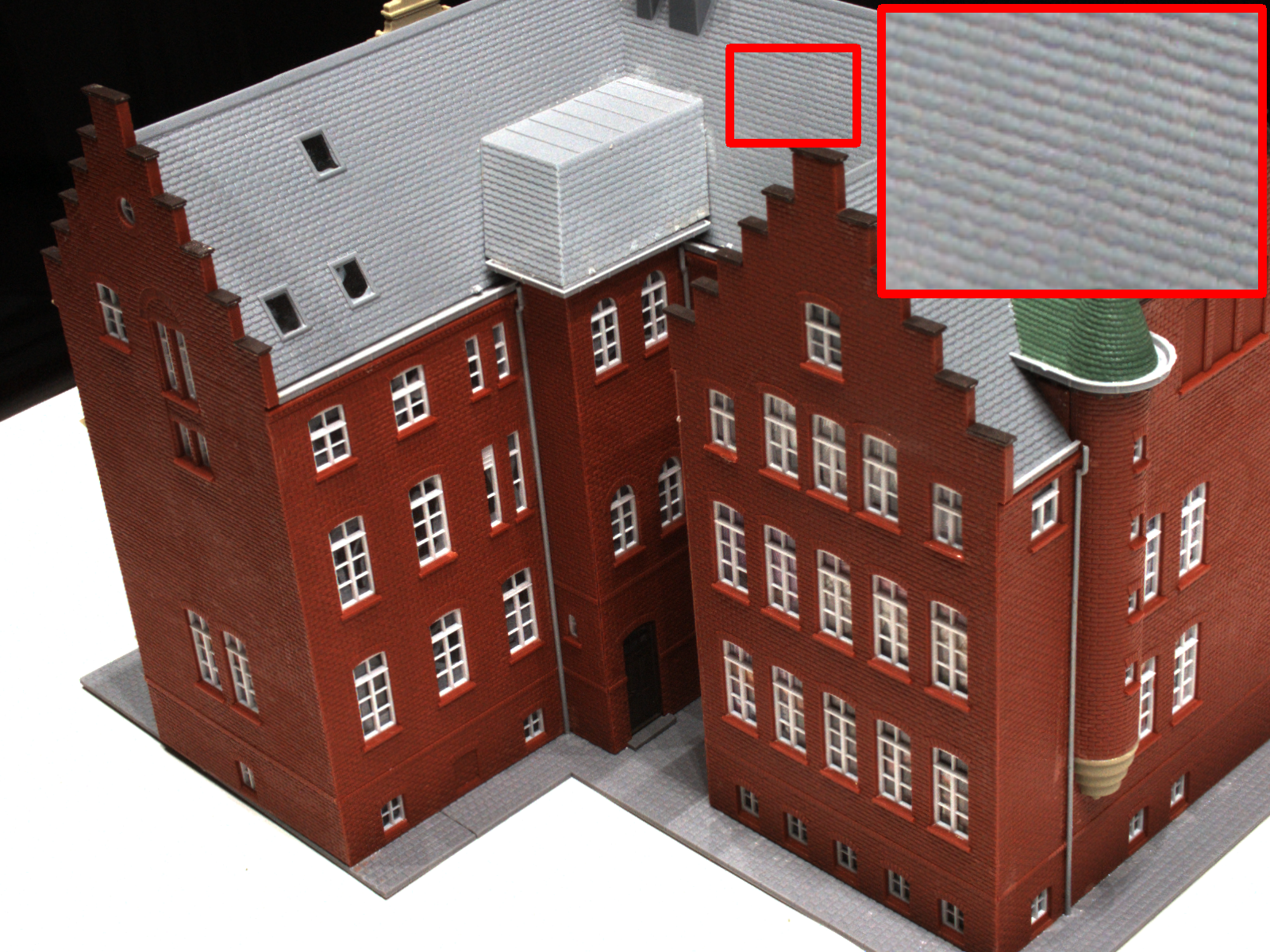}}
  \mpage{0.22}{\includegraphics[width=\linewidth]{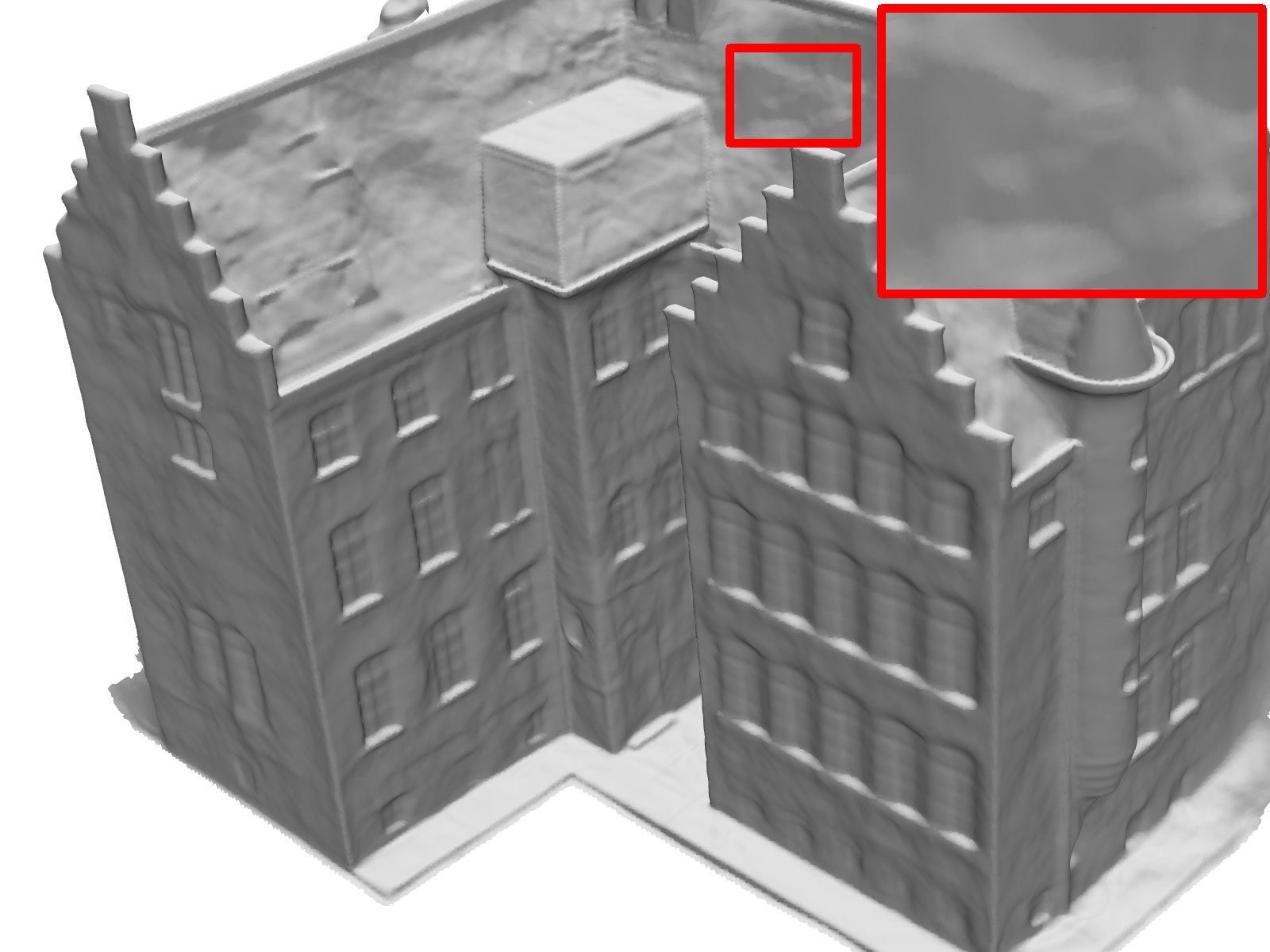}}
  \mpage{0.22}{\includegraphics[width=\linewidth]{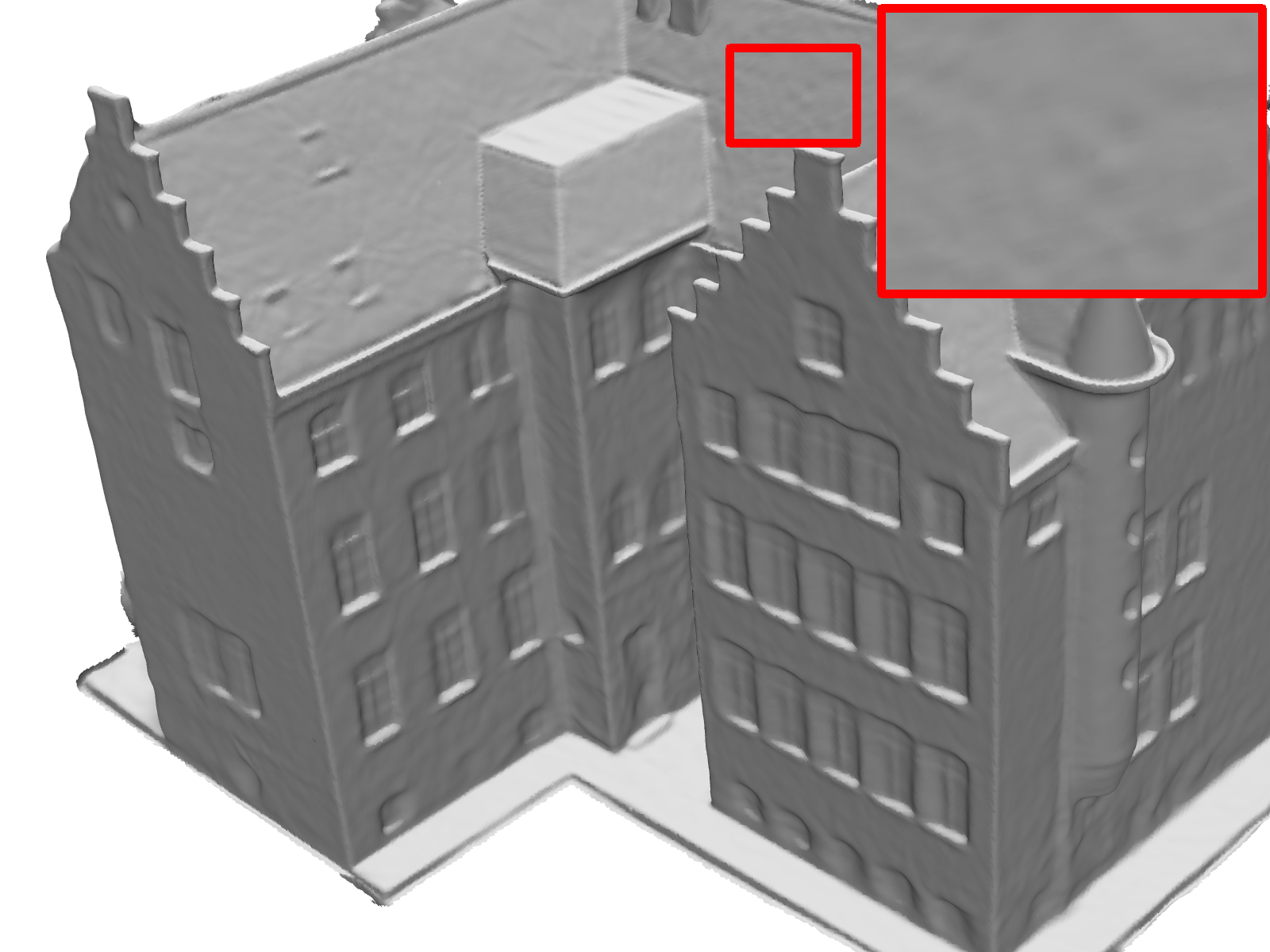}}
  \mpage{0.22}{\includegraphics[width=\linewidth]{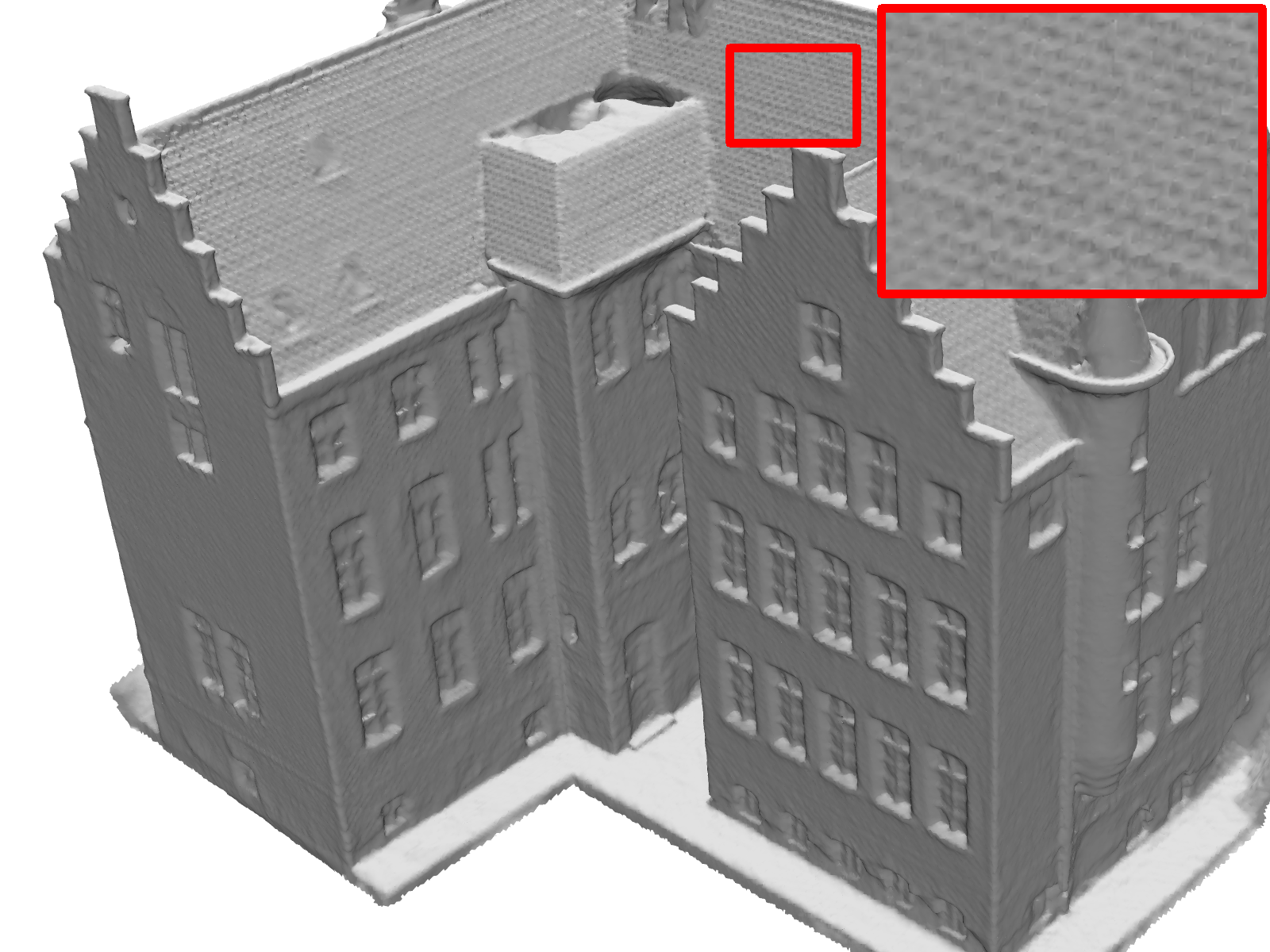}}
   \\
 \rotatebox[origin=C]{90}{\parbox{20mm}{\centering \small Scan 55 \\ (DTU)}} 
  \mpage{0.22}{\includegraphics[width=\linewidth]{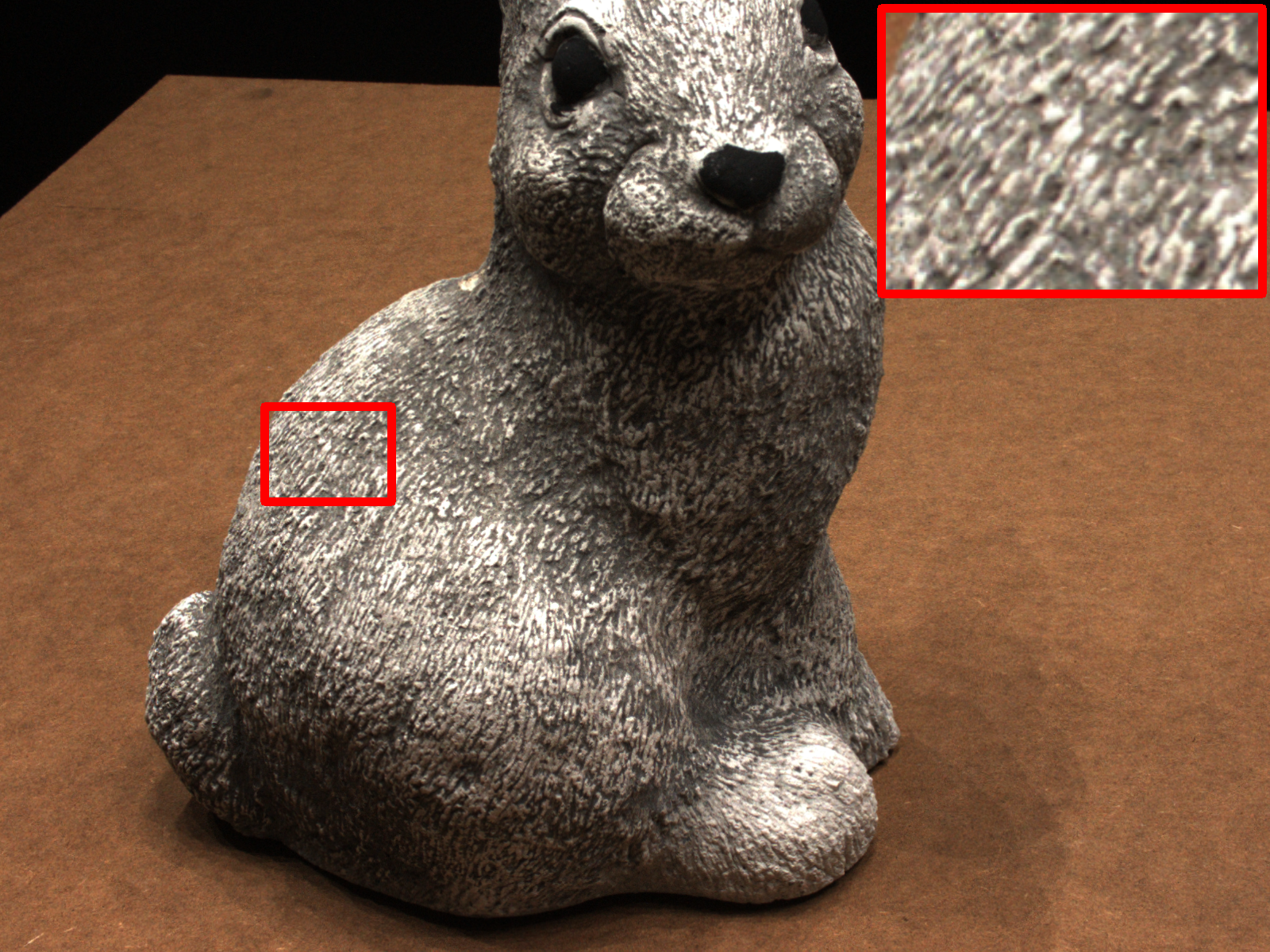}}
  \mpage{0.22}{\includegraphics[width=\linewidth]{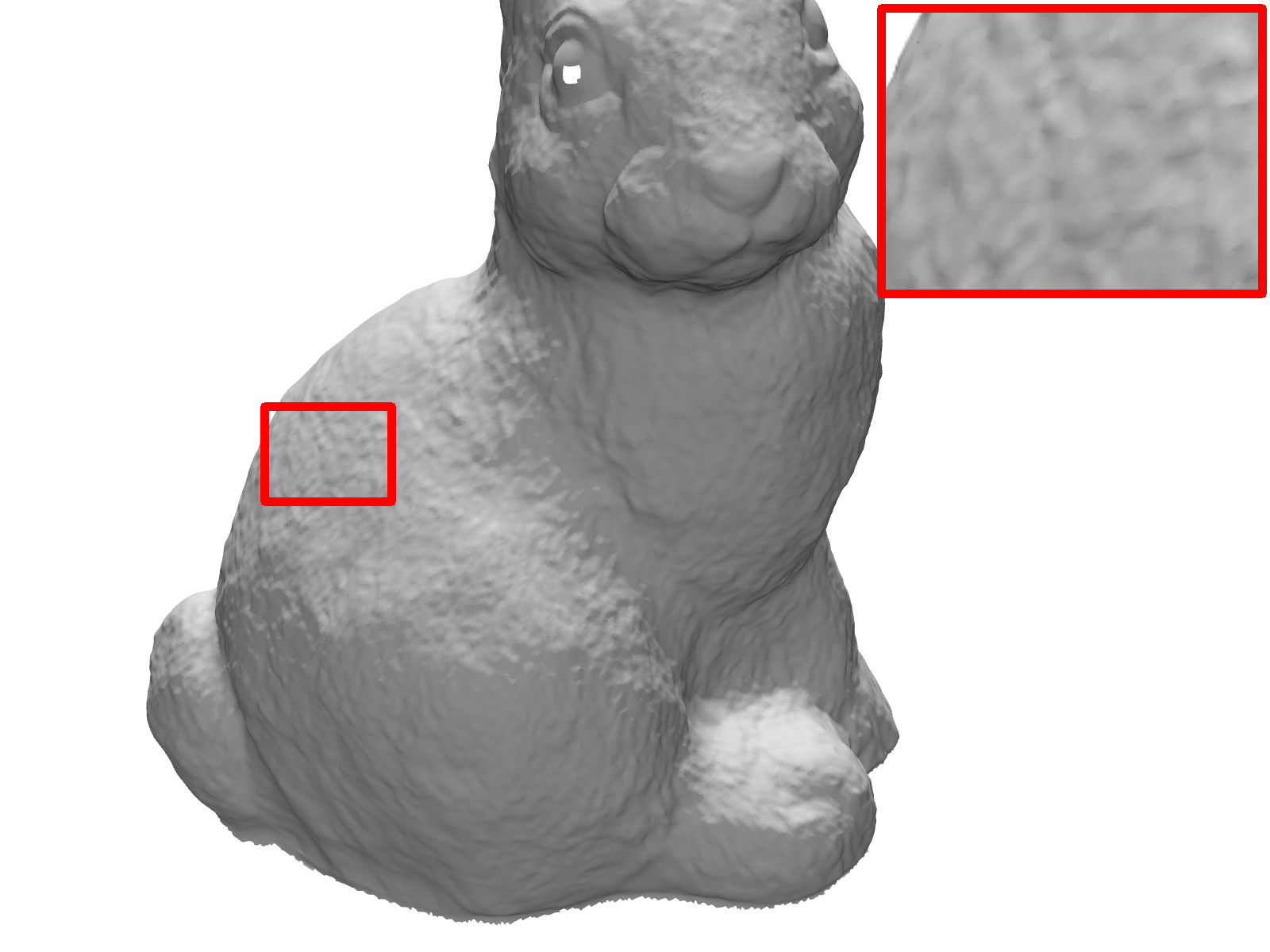}}
  \mpage{0.22}{\includegraphics[width=\linewidth]{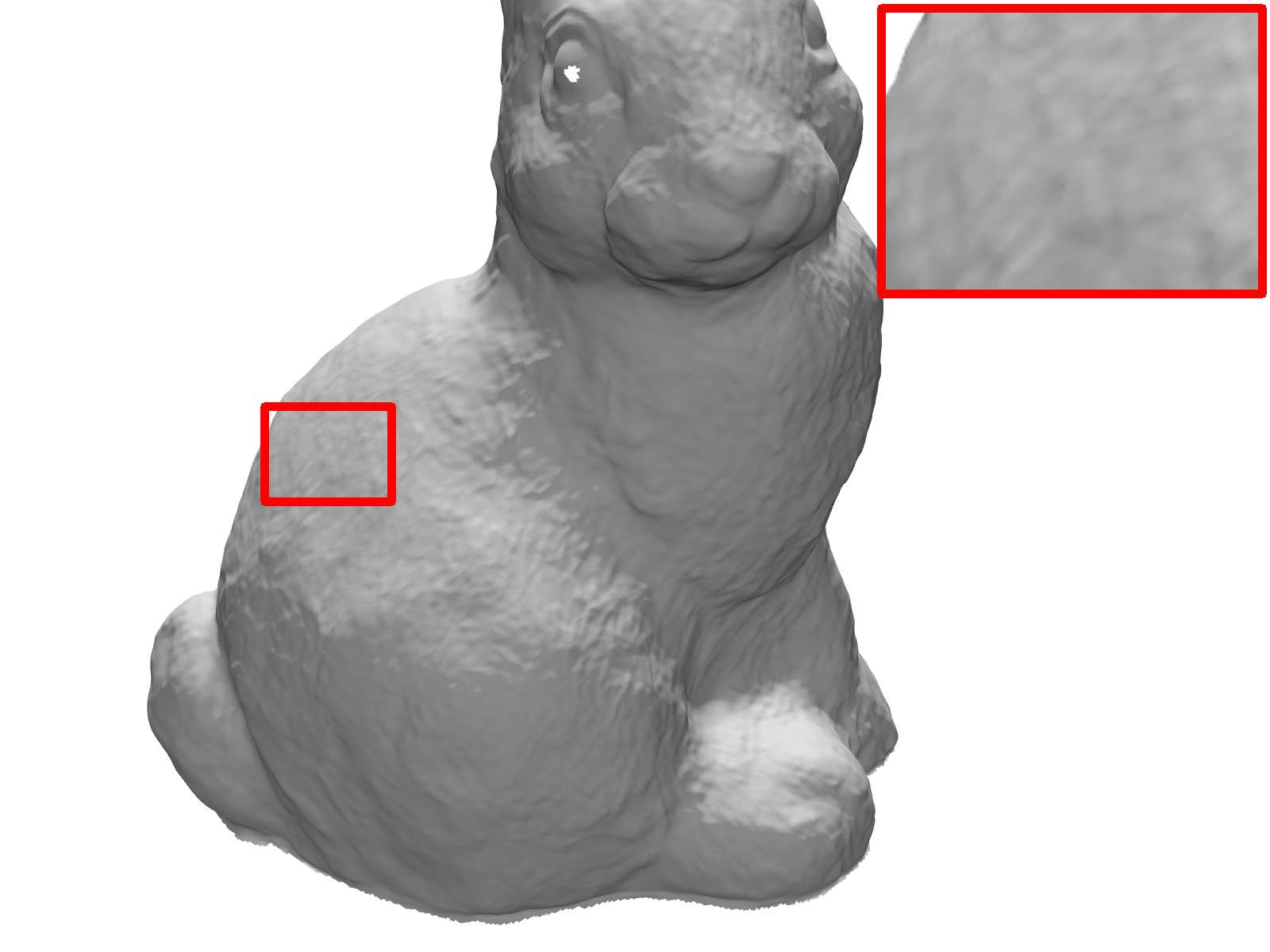}}
  \mpage{0.22}{\includegraphics[width=\linewidth]{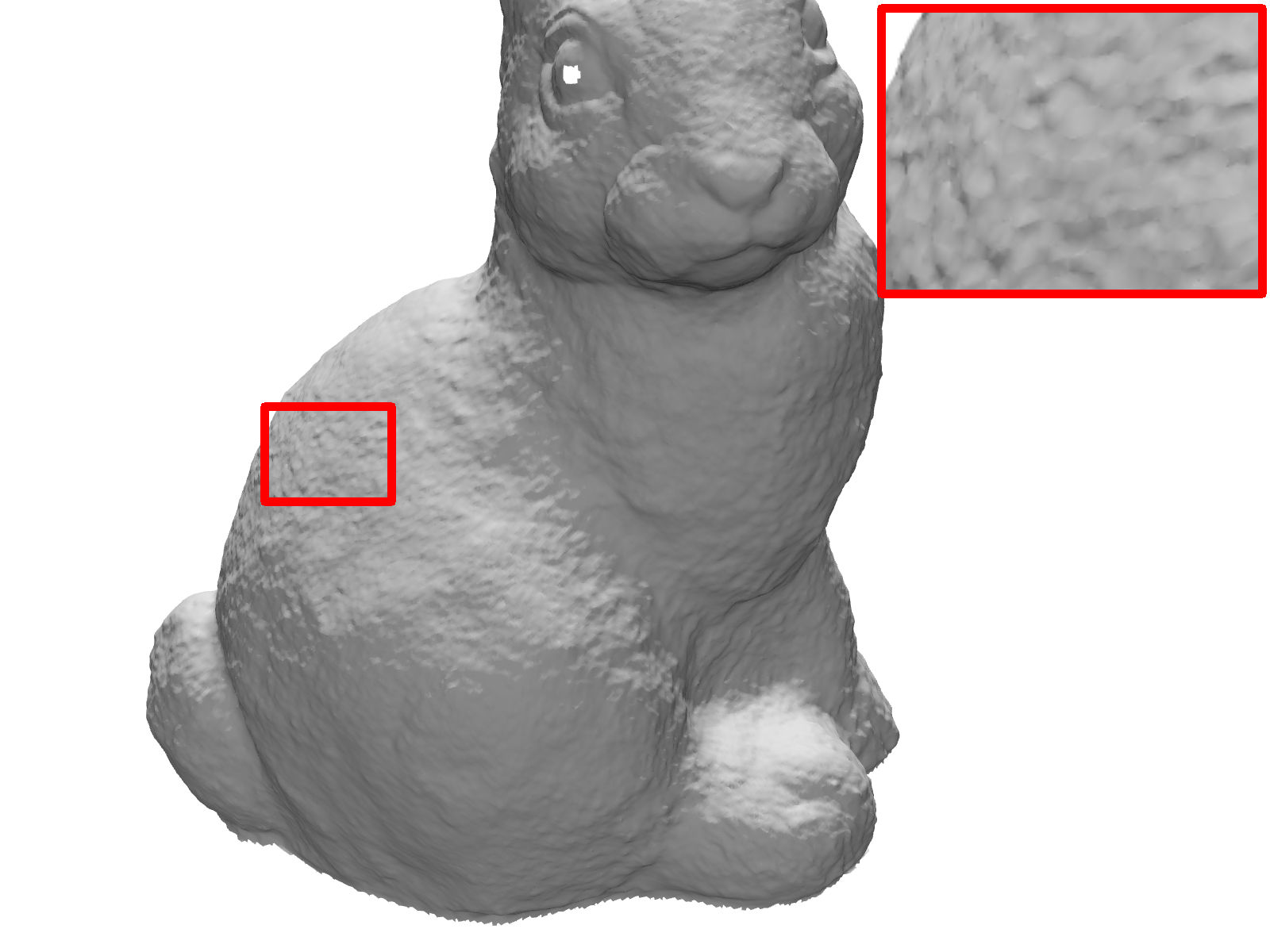}}
  \\
 \rotatebox[origin=C]{90}{\parbox{20mm}{\centering \small Scan 63 \\ (DTU)}} 
  \mpage{0.22}{\includegraphics[width=\linewidth]{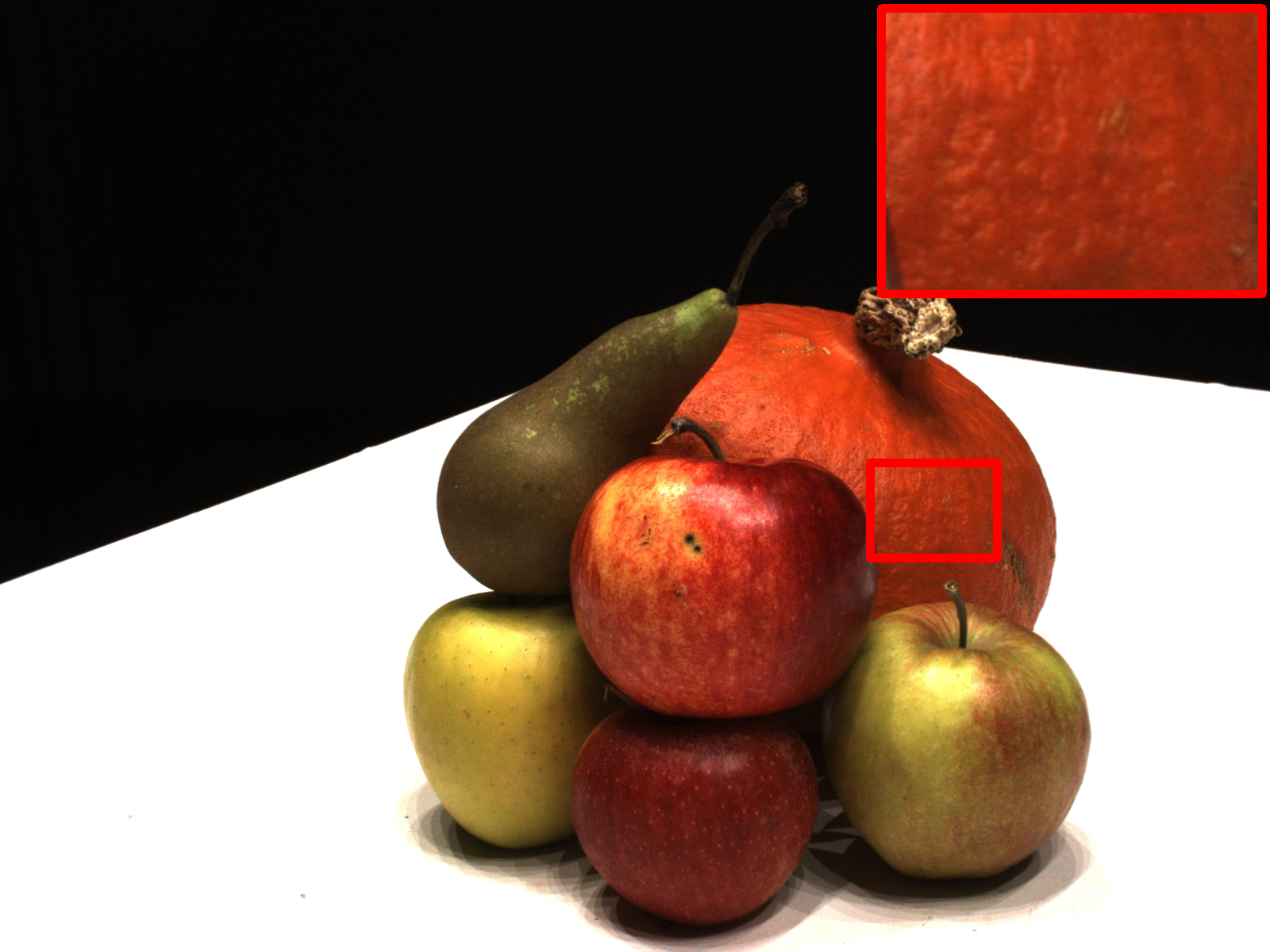}}
  \mpage{0.22}{\includegraphics[width=\linewidth]{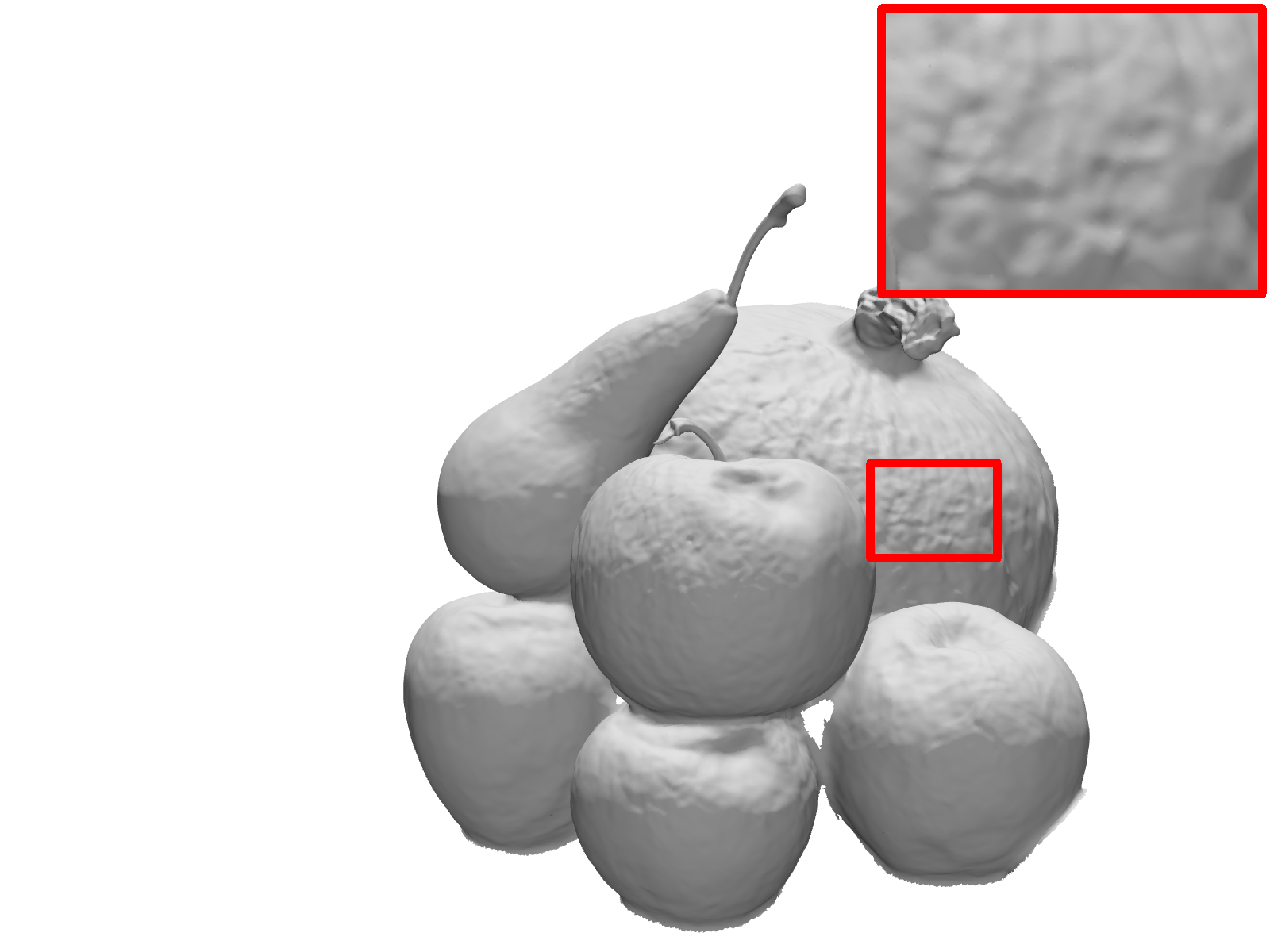}}
  \mpage{0.22}{\includegraphics[width=\linewidth]{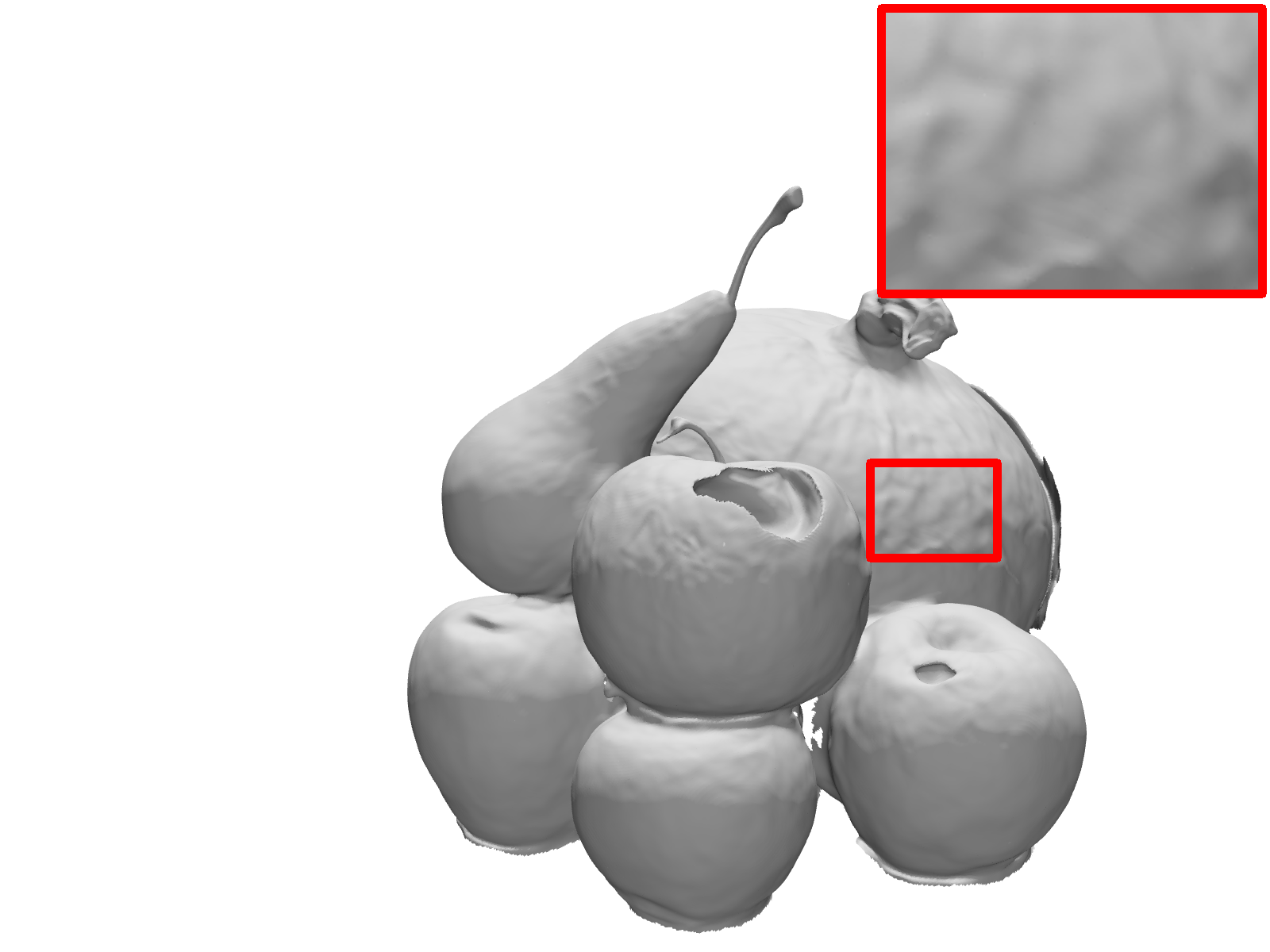}}
  \mpage{0.22}{\includegraphics[width=\linewidth]{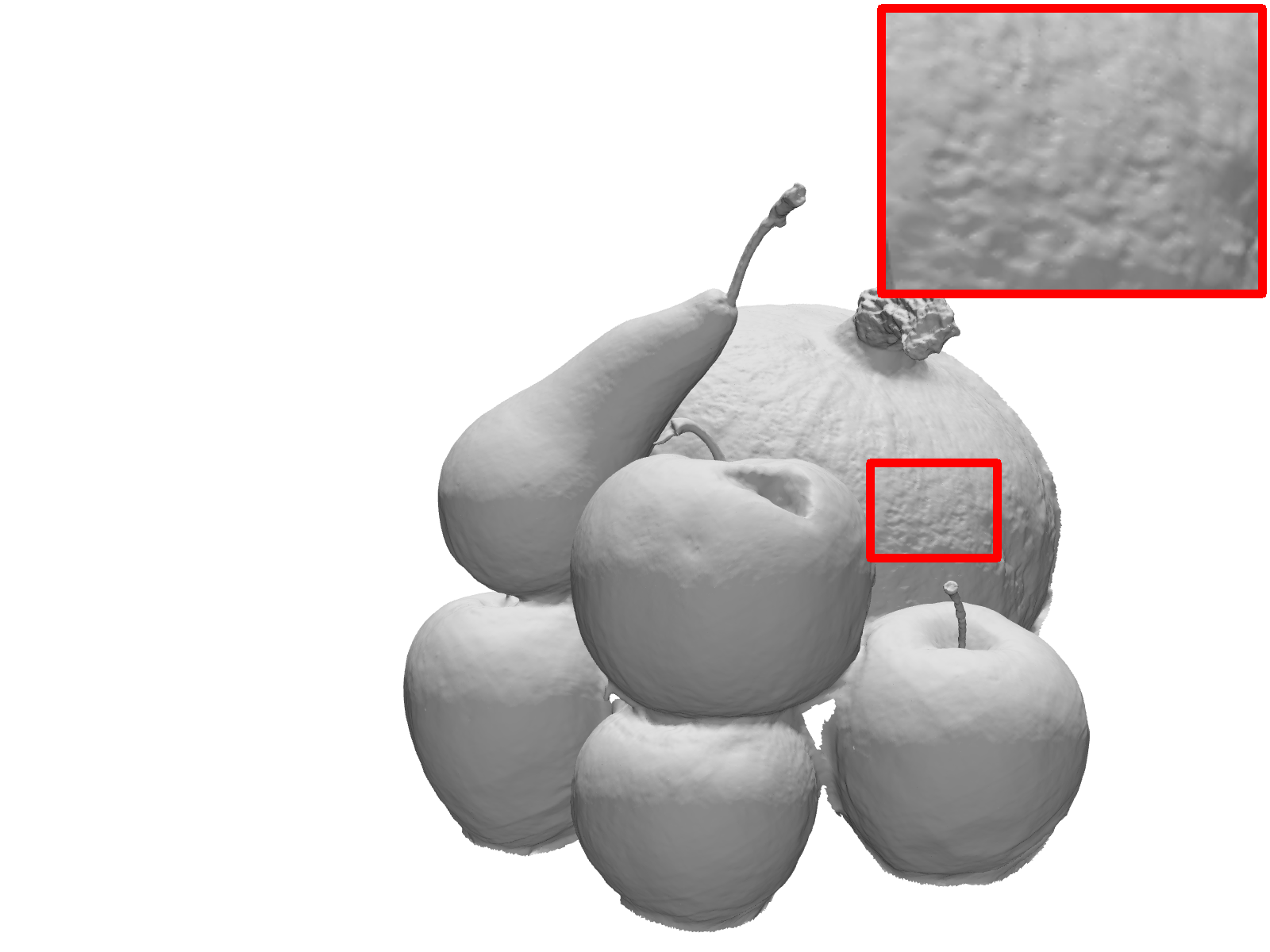}}
  \\
 \rotatebox[origin=C]{90}{\parbox{20mm}{\centering \small Scan 69 \\ (DTU)}} 
  \mpage{0.22}{\includegraphics[width=\linewidth]{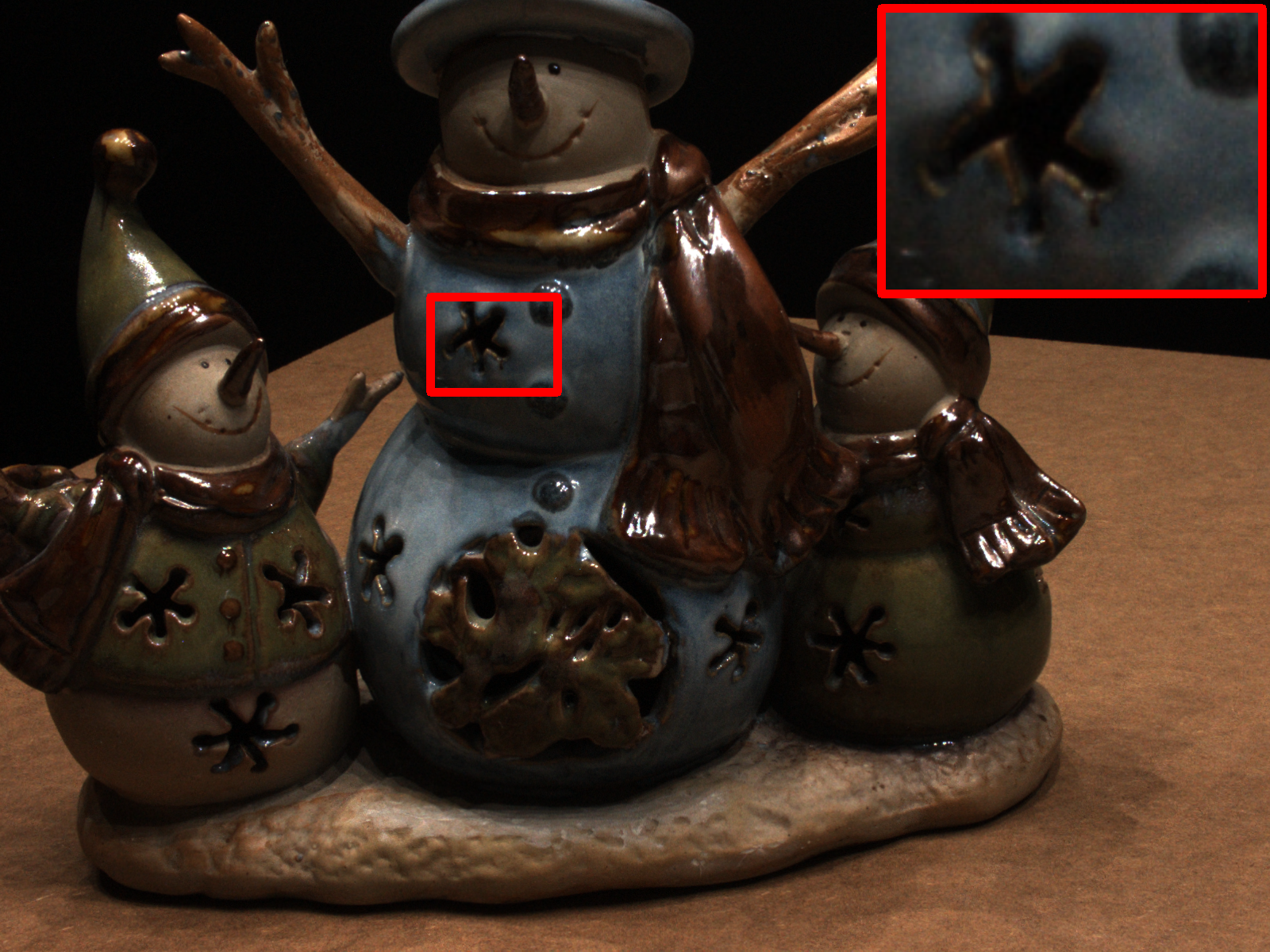}}
  \mpage{0.22}{\includegraphics[width=\linewidth]{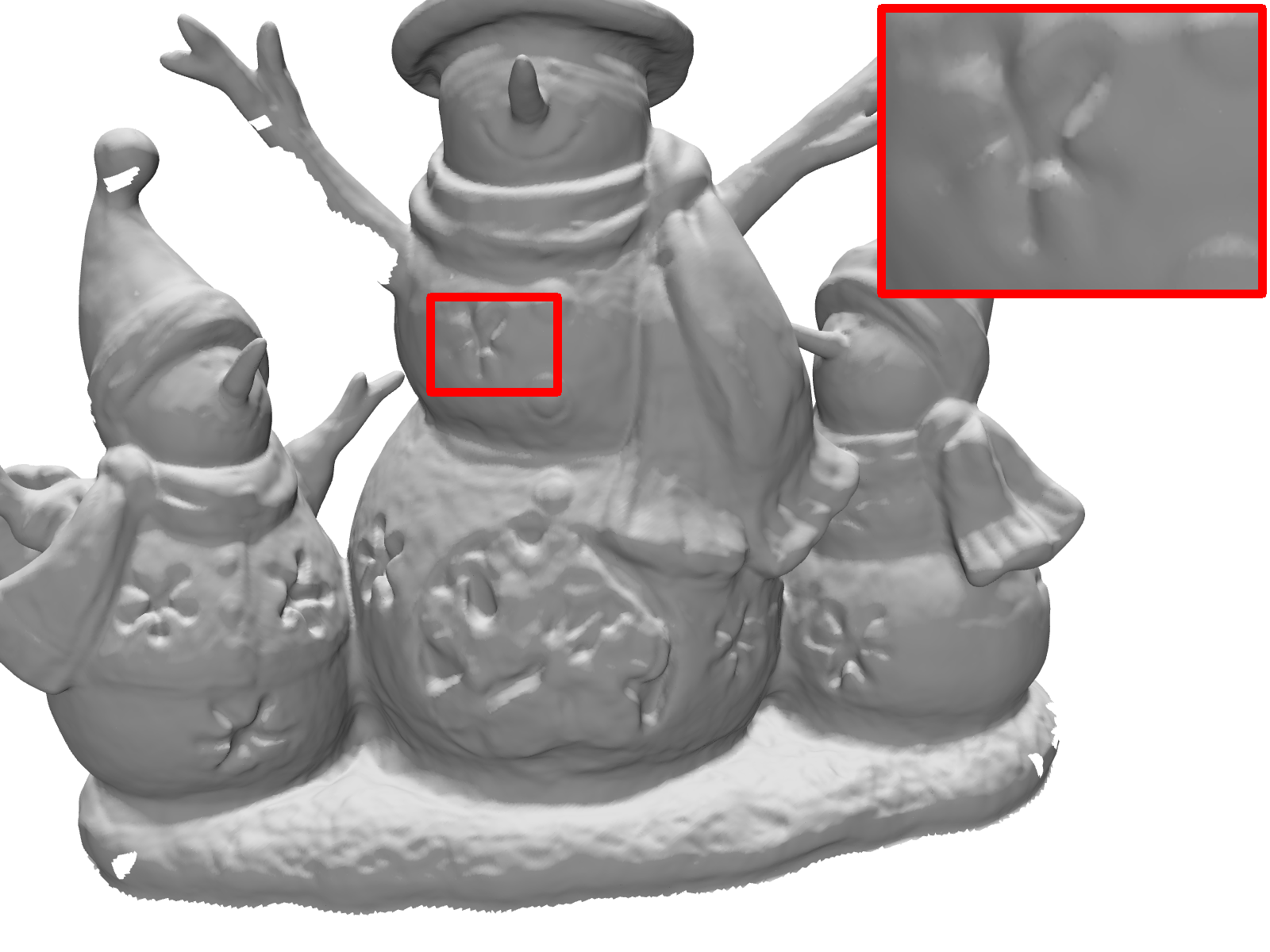}}
  \mpage{0.22}{\includegraphics[width=\linewidth]{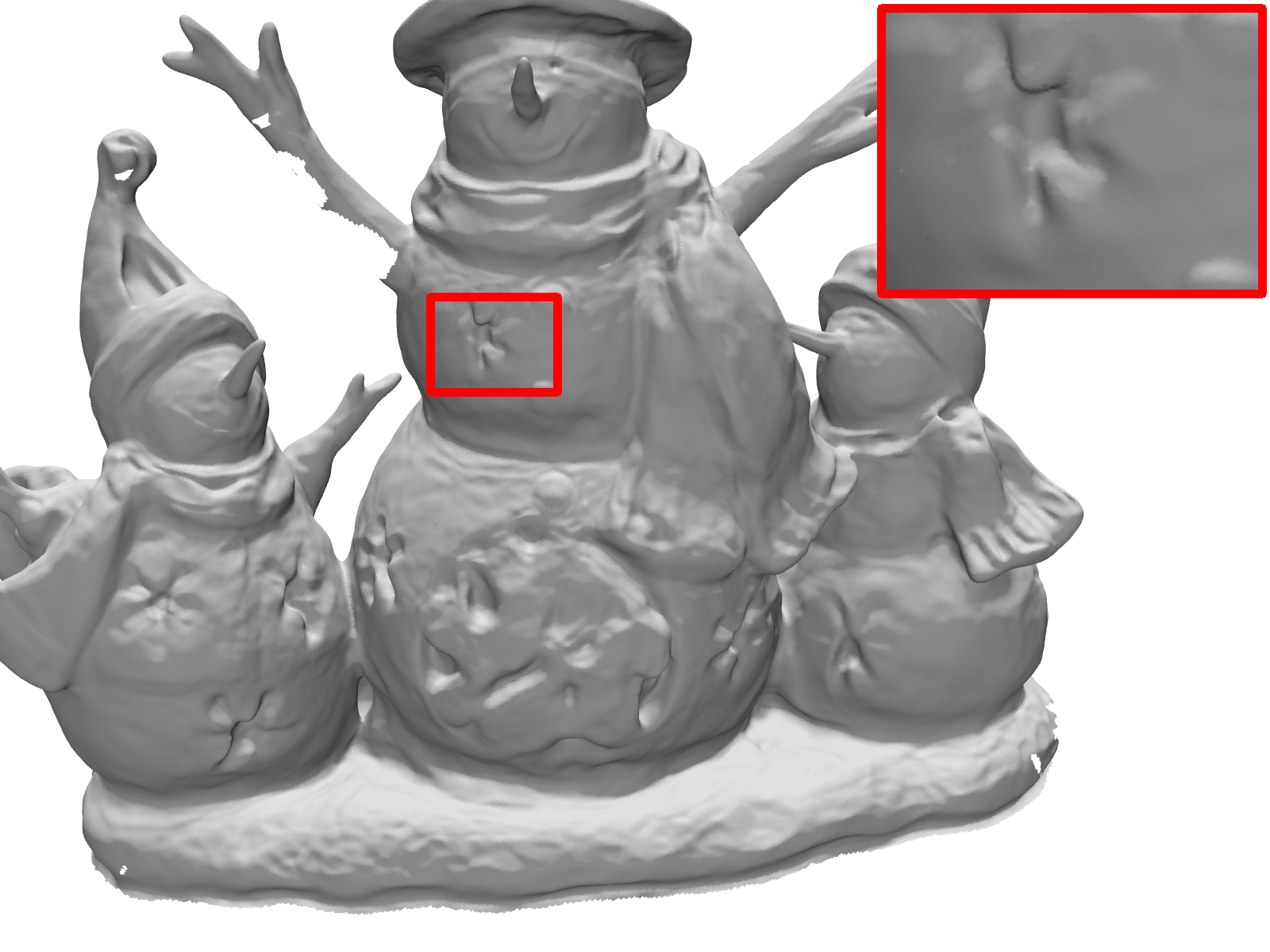}}
  \mpage{0.22}{\includegraphics[width=\linewidth]{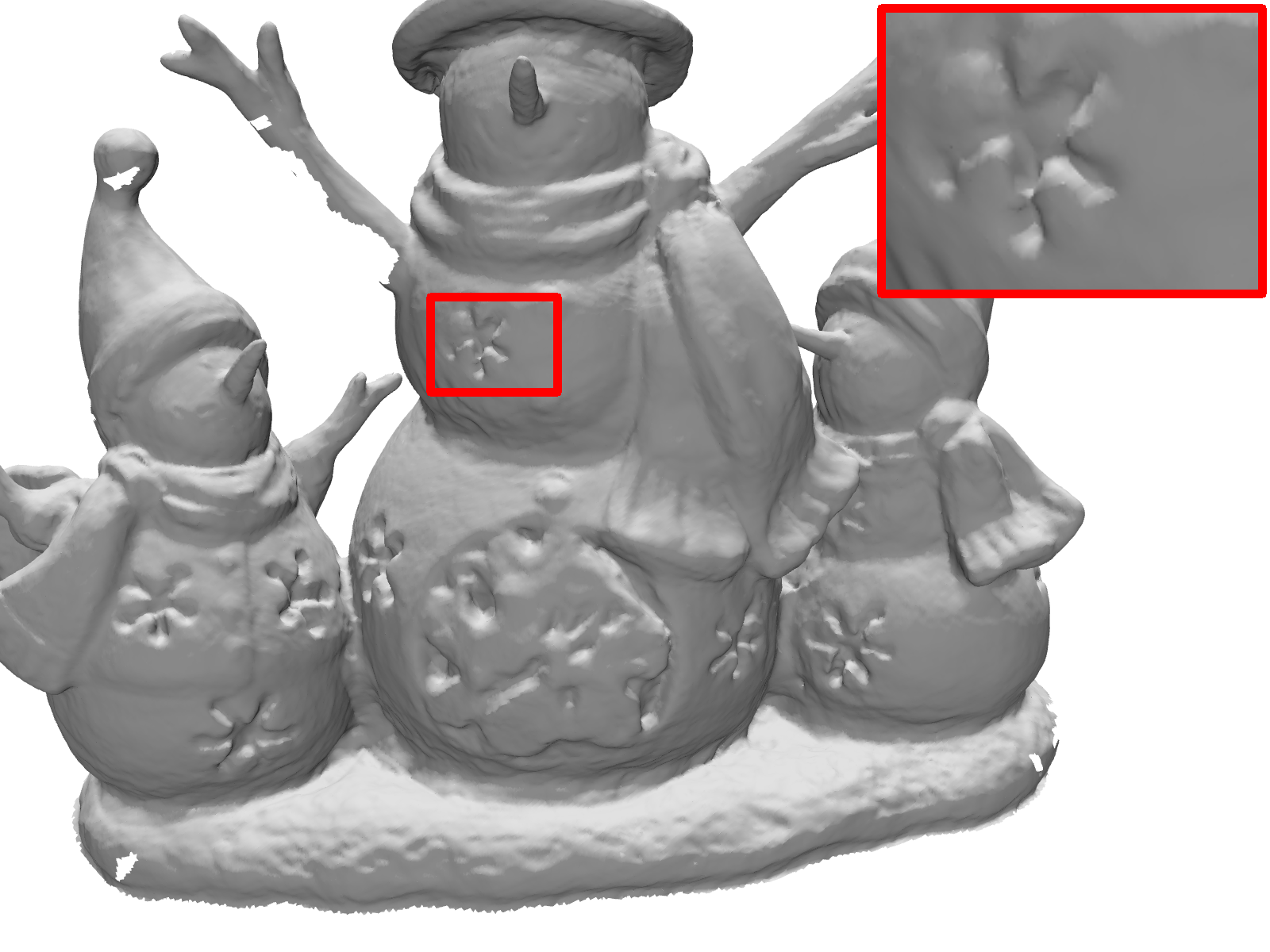}}
   \\
 \rotatebox[origin=C]{90}{\parbox{20mm}{\centering \small Scan 118 \\ (DTU)}} 
  \mpage{0.22}{\includegraphics[width=\linewidth]{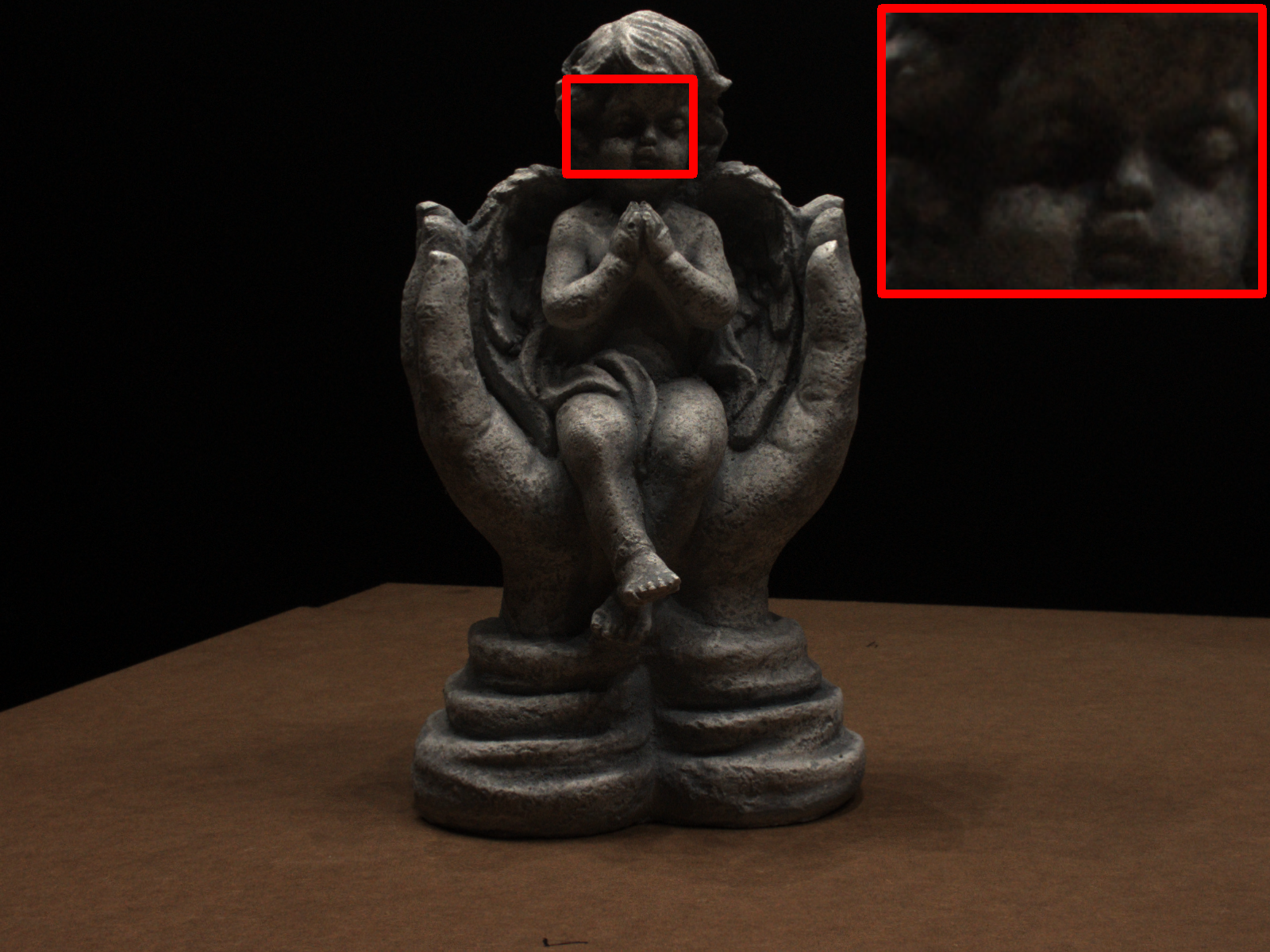}}
  \mpage{0.22}{\includegraphics[width=\linewidth]{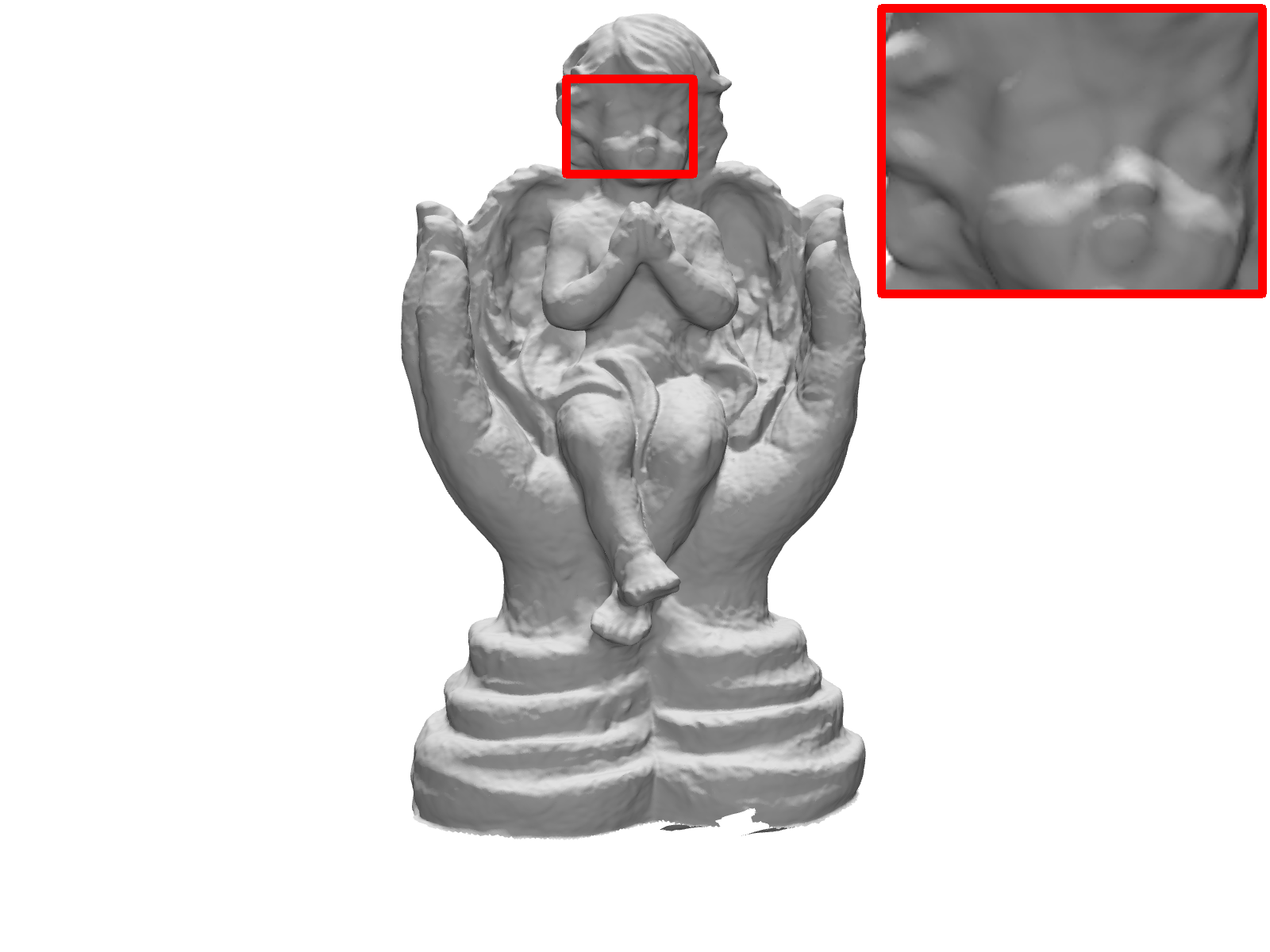}}
  \mpage{0.22}{\includegraphics[width=\linewidth]{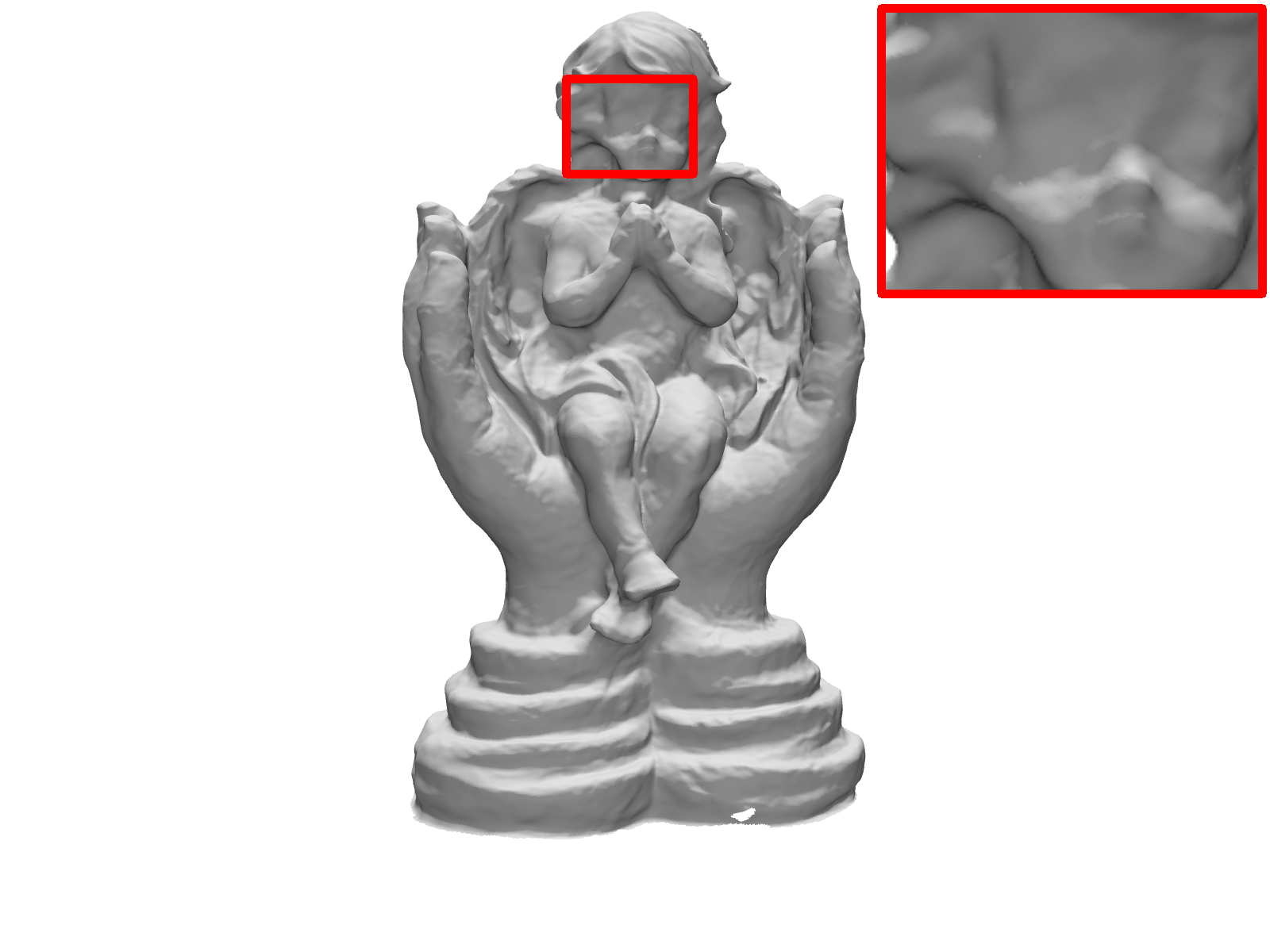}}
  \mpage{0.22}{\includegraphics[width=\linewidth]{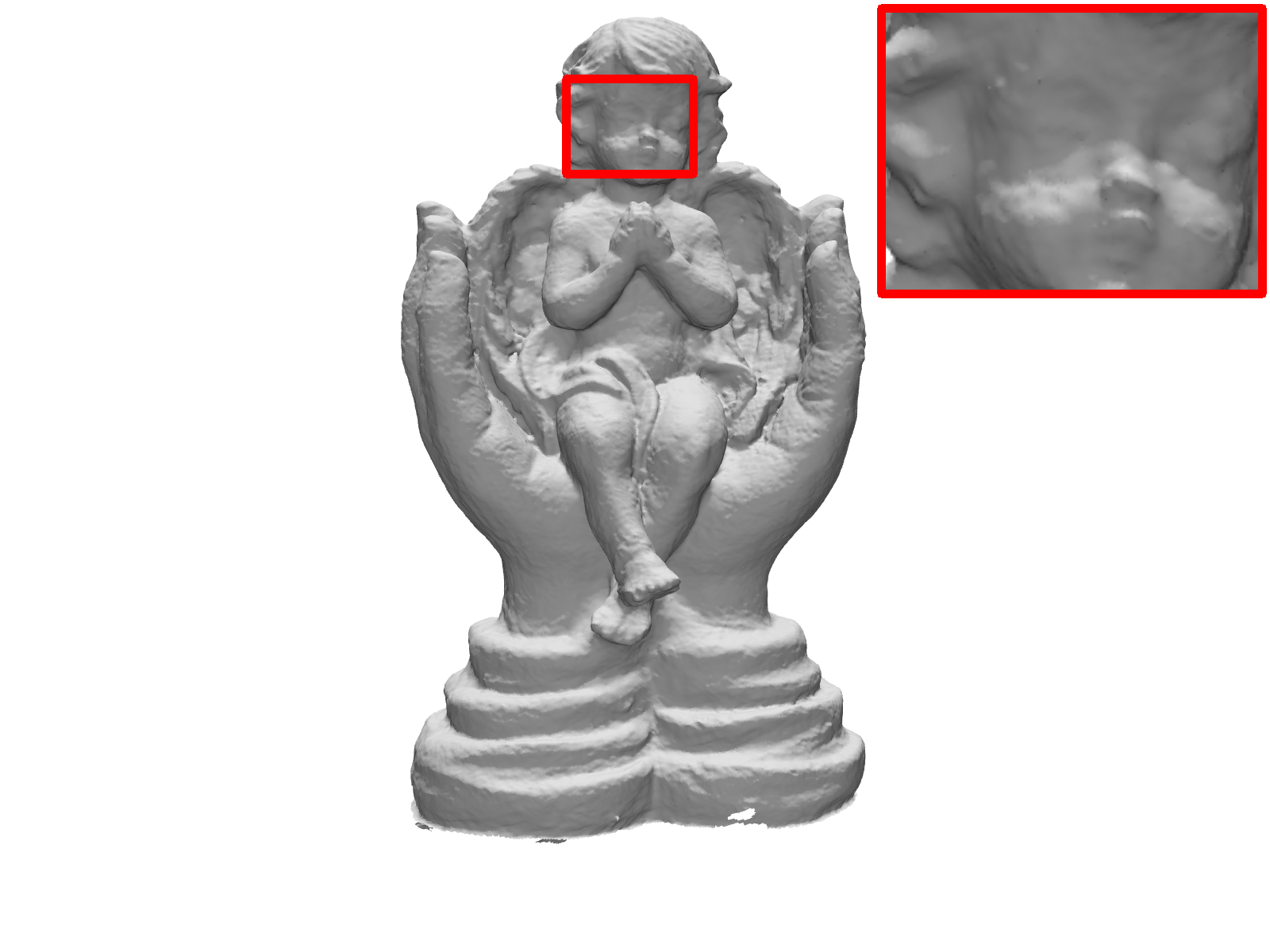}}
   \\
  \mpage{0.05}{\ }
  \mpage{0.215}{Reference Image}
  \mpage{0.215}{NeuS}
  \mpage{0.215}{NeuralWarp}
  \mpage{0.215}{Ours}
  
   \caption{Qualitative Results on DTU\cite{dtu} dataset. Our work retains high frequency details of the surface while preserving overall geometric correctness.}
   \vspace{-5pt}
\label{fig:dtu_qual}
\end{figure*}

\begin{figure*}[!h]
 \centering
 \rotatebox[origin=C]{90}{\parbox{20mm}{\centering \small Clock \\ (BlendedMVS)}} 
  \mpage{0.30}{\includegraphics[width=\linewidth]{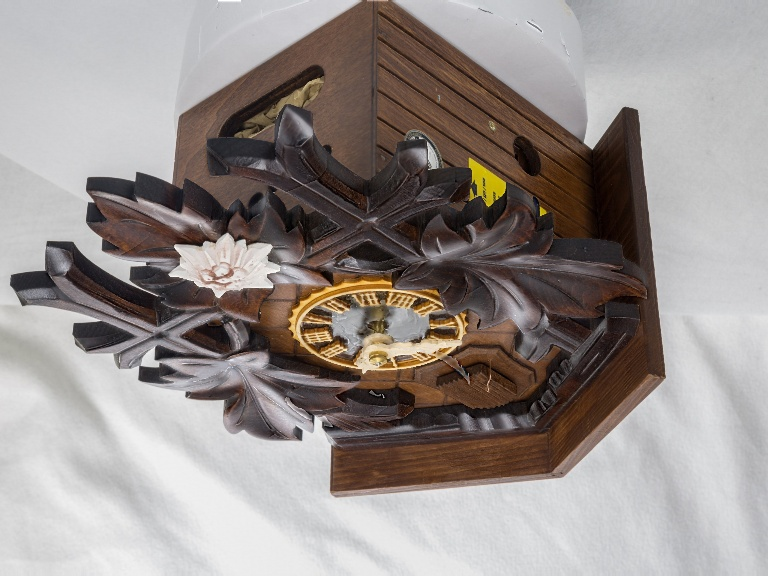}}
  \mpage{0.30}{\includegraphics[width=\linewidth]{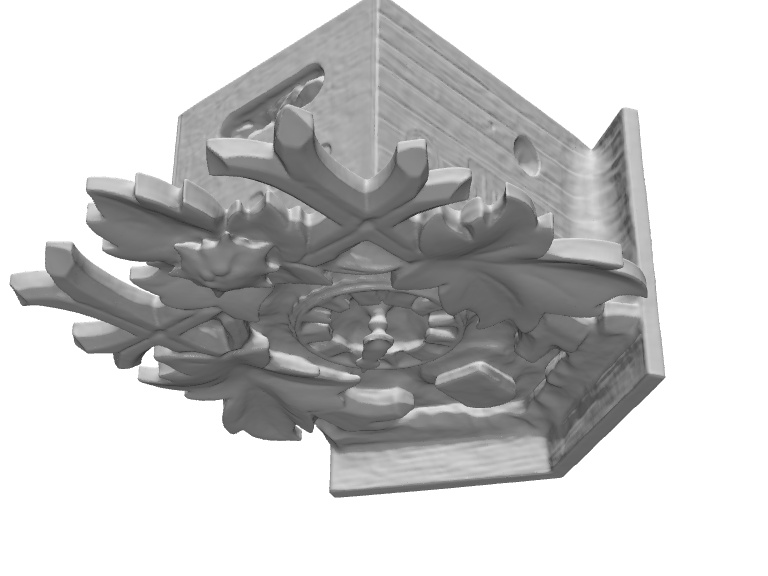}}
  \mpage{0.30}{\includegraphics[width=\linewidth]{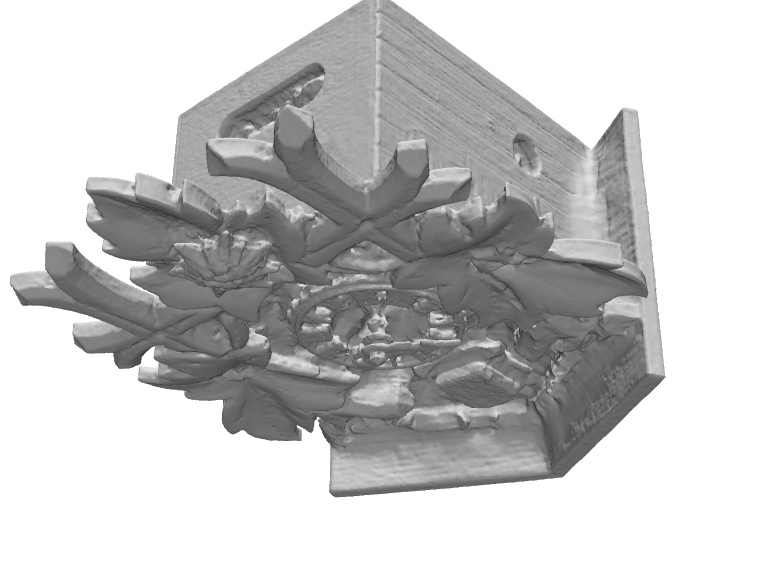}}
   \\
 \rotatebox[origin=C]{90}{\parbox{20mm}{\centering \small Durian \\ (BlendedMVS)}} 
  \mpage{0.30}{\includegraphics[width=\linewidth]{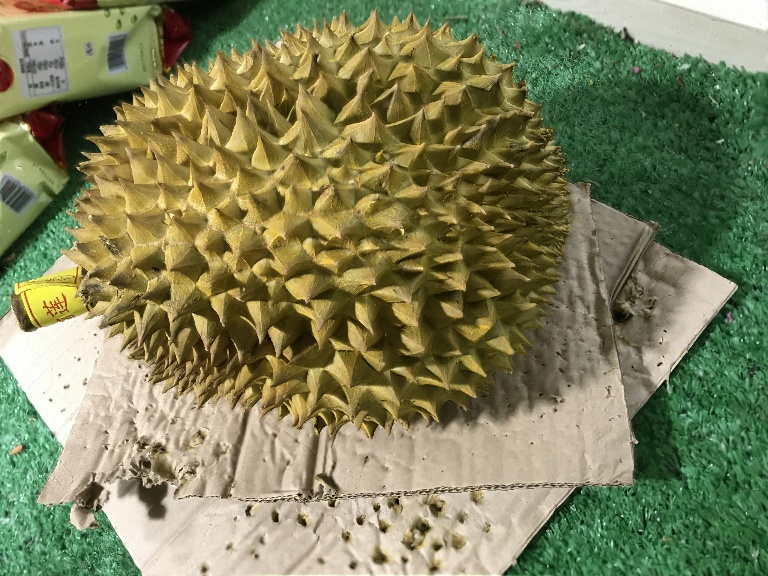}}
  \mpage{0.30}{\includegraphics[width=\linewidth]{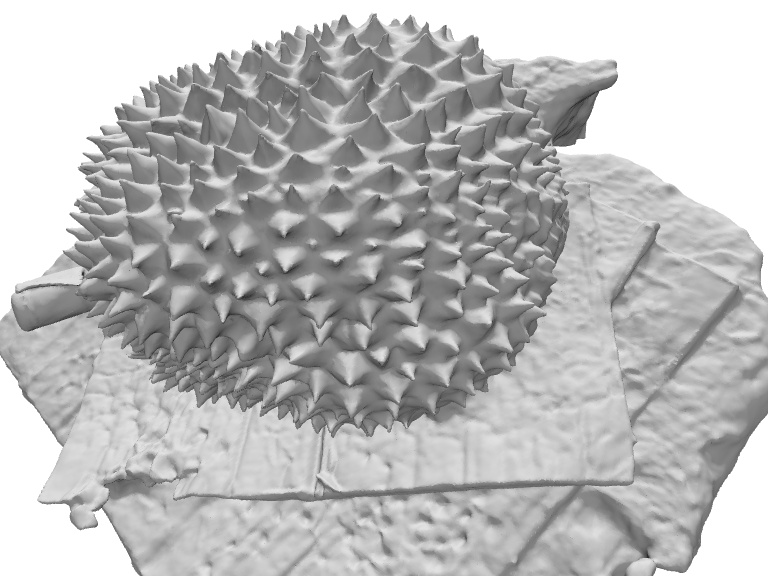}}
  \mpage{0.30}{\includegraphics[width=\linewidth]{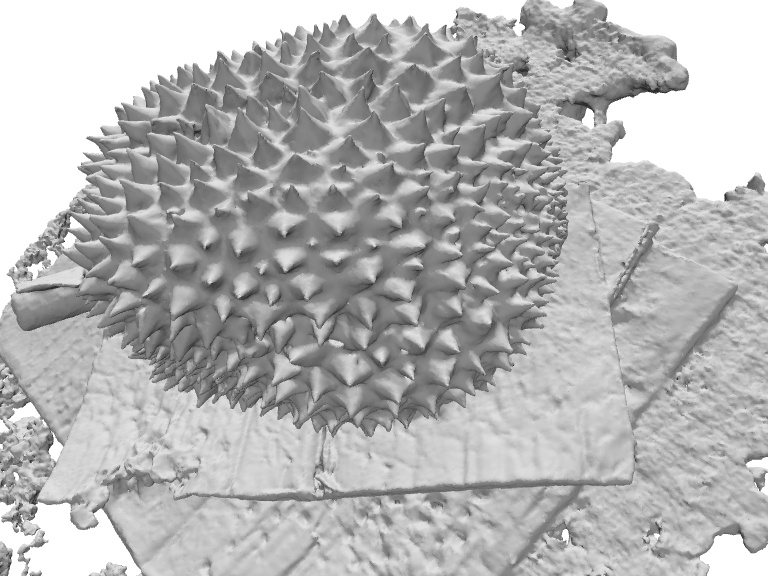}}
  \\
  \mpage{0.05}{\ }
  \mpage{0.30}{Reference Image}
  \mpage{0.30}{NeuS}
  \mpage{0.30}{Ours}
  
   \caption{Qualitative Results on BlendedMVS\cite{blendedmvs} dataset. Our work retains high frequency details of the surface while preserving overall geometric correctness. Notice the clock face and spikes on durian are more detailed}
   
\label{fig:dtu_qual_bmvs}
\end{figure*}

\section{Experiments}
\label{sec:experiments}
\subsection{Experiment Setup}
\noindent \textbf{Datasets.} To evaluate the results of our method, we run our model on the same 15 scenes on DTU \cite{dtu} datasets, following previous works \cite{idr, neus, geoneus, francois2022neuralwarp}. Each scene contains 49 to 64 images of size 1600 $\times$ 1200, with various materials and complex geometries. Each scene also contains annotated foreground masks from \cite{idr}, which we do not utilize in our algorithm. We also use BlendedMVS~\cite{blendedmvs} for qualitative comparison.

\noindent \textbf{Baselines.} For baselines, we evaluate against NeuS\cite{neus} when running without segmentation masks, and recent work NeuralWarp \cite{francois2022neuralwarp}. We use marching cubes \cite{mcubes} with 512 resolution to extract the final mesh. To make a fair comparison, we apply the same post-processing step on all models. Specifically, we follow the mesh filtering implementation using visibility masks dilated by 12 pixels from \cite{francois2022neuralwarp}. We then take the largest connected component by area, and finally prune triangles that are never seen in any of the original input views.

\subsection{Comparison}

We evaluate the quality of the macro geometry our algorithm recovers through quantitative comparisons with other algorithms in the setting of surface reconstruction without mask supervision. Overall reconstruction accuracy is measured using Chamfer distances, which is calculated in the same way that NeuS \cite{neus} and NeuralWarp \cite{francois2022neuralwarp} do, and we report these metrics in Table \ref{tab:dtu_result}. Overall, our reconstructions are more accurate compared to NeuS, and are within the same ballpark of NeuralWarp's results.

However, an algorithm's ability to recover high frequency details cannot be reflected through Chamfer distance, so we resort to qualitative evaluation instead. In Figure \ref{fig:dtu_qual}, we provide side by side comparisons of reconstructed surfaces produced by NeuS and our algorithm, as well as original reference images from the DTU dataset at which the snapshot of the surface is captured. Overall, our surfaces exhibit much finer details and texturing compared to NeuS. In Scan 24, our method is the only one that can preserve the small tiles bump on the roof, and it also has sharp window frames on the building. We do also observe some artifacts such as a new hole/dent appearing on the roof in Scan 24 and on the apple in Scan 63. We also provide qualitative results on 2 scenes from BlendedMVS\cite{blendedmvs} dataset. See Figure~\ref{fig:dtu_qual_bmvs}
 
\subsection{Ablation Study}
\begin{figure*}[!h]
 \centering
 \rotatebox[origin=C]{90}{\parbox{20mm}{\centering \small Scan 24 \\ (DTU)}} 
  \mpage{0.22}{\includegraphics[width=\linewidth]{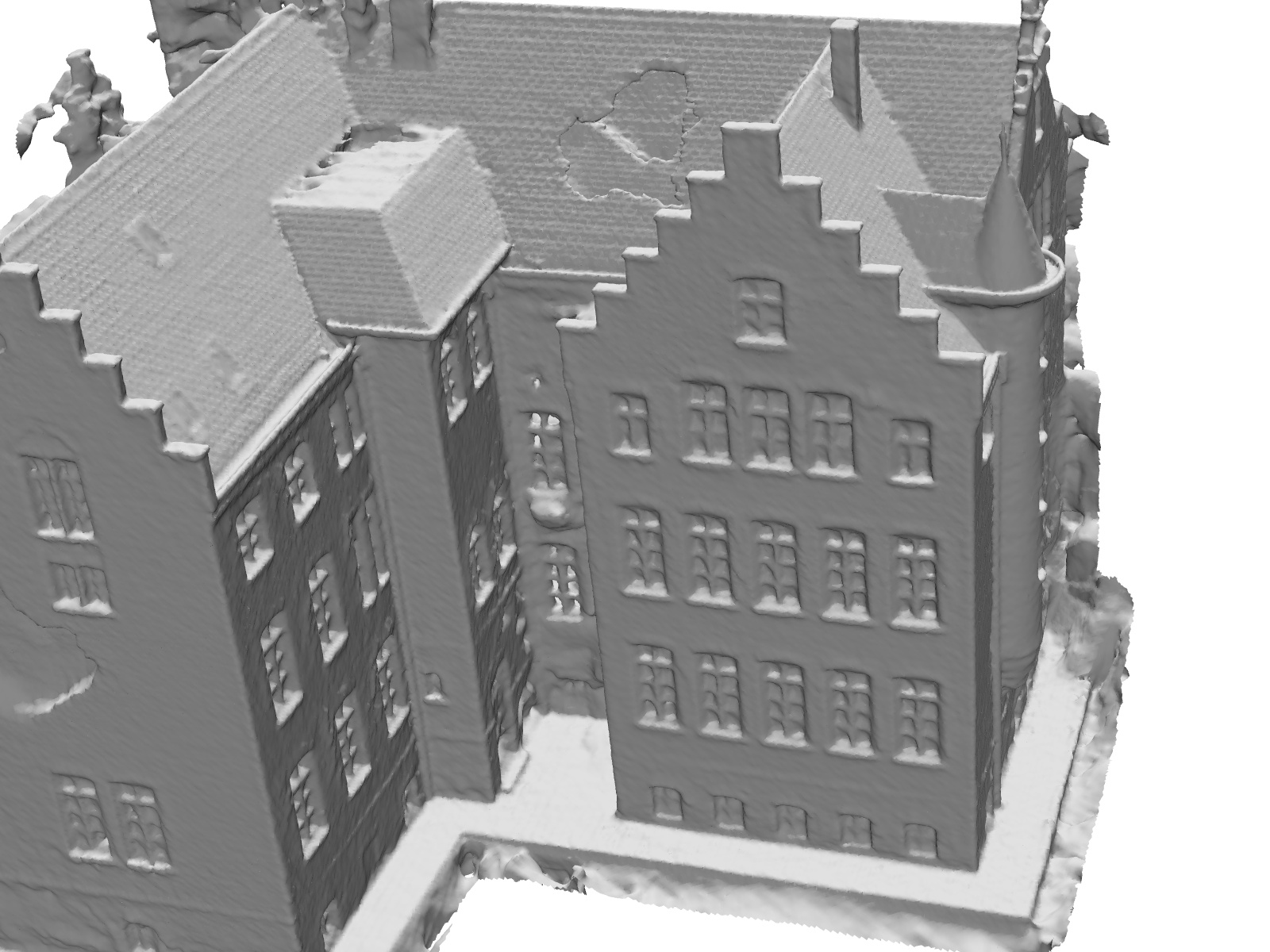}
  \caption*{Trained Directly}}
  \mpage{0.22}{\includegraphics[width=\linewidth]{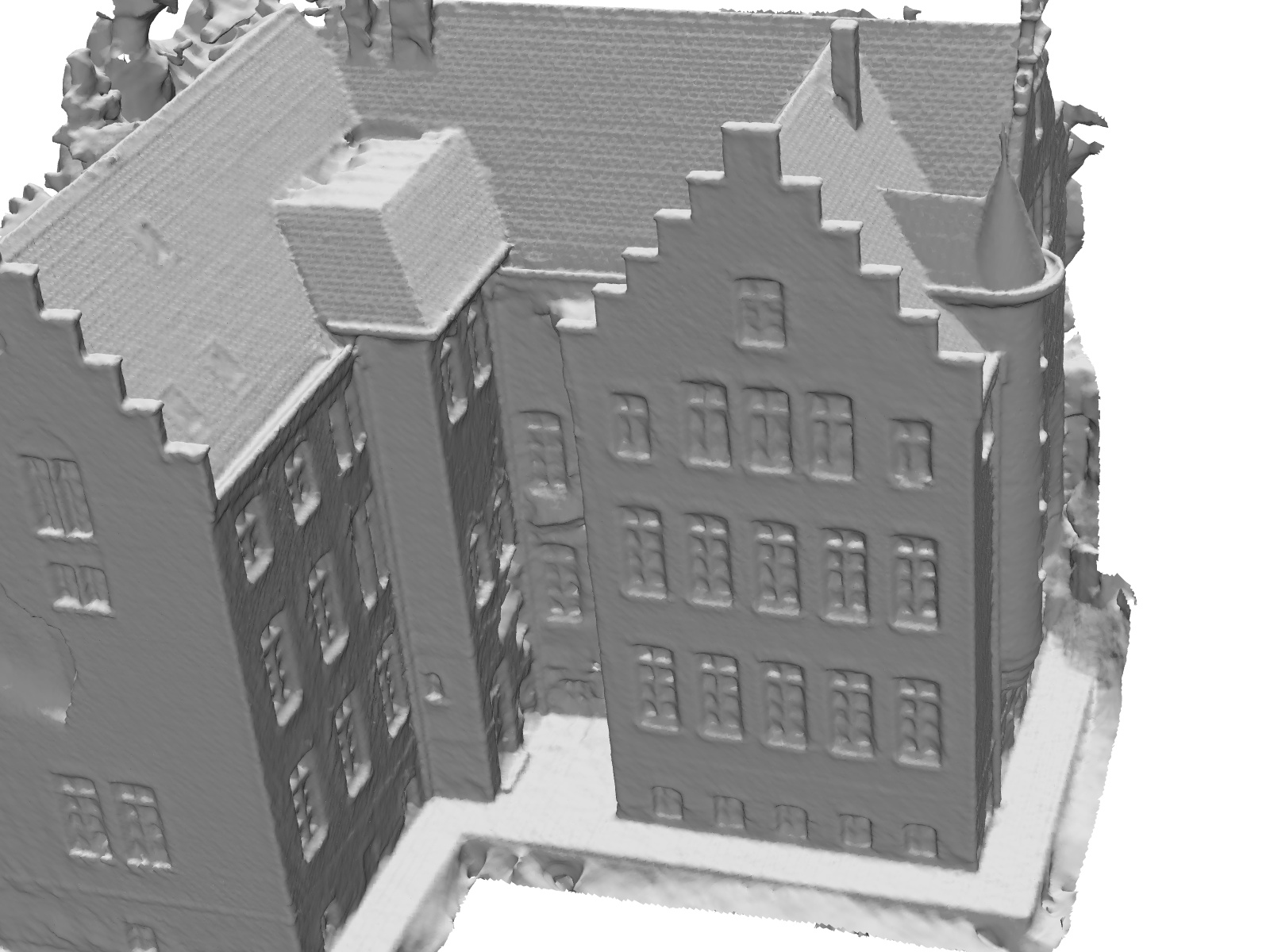}
  \caption*{No coarse-to-fine annealing}}
  \mpage{0.22}{\includegraphics[width=\linewidth]{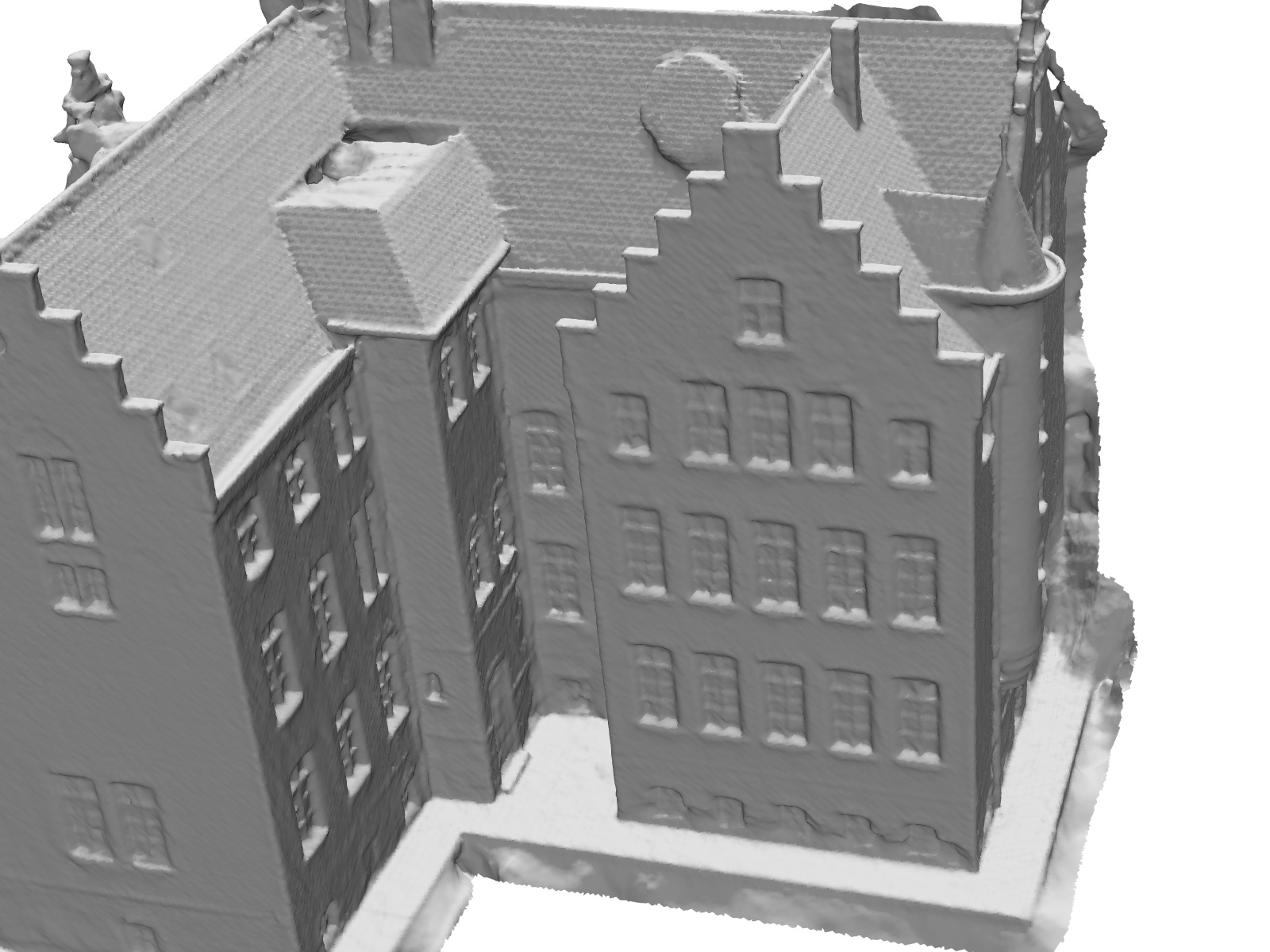}
  \caption*{No regularization}}
  \mpage{0.22}{\includegraphics[width=\linewidth]{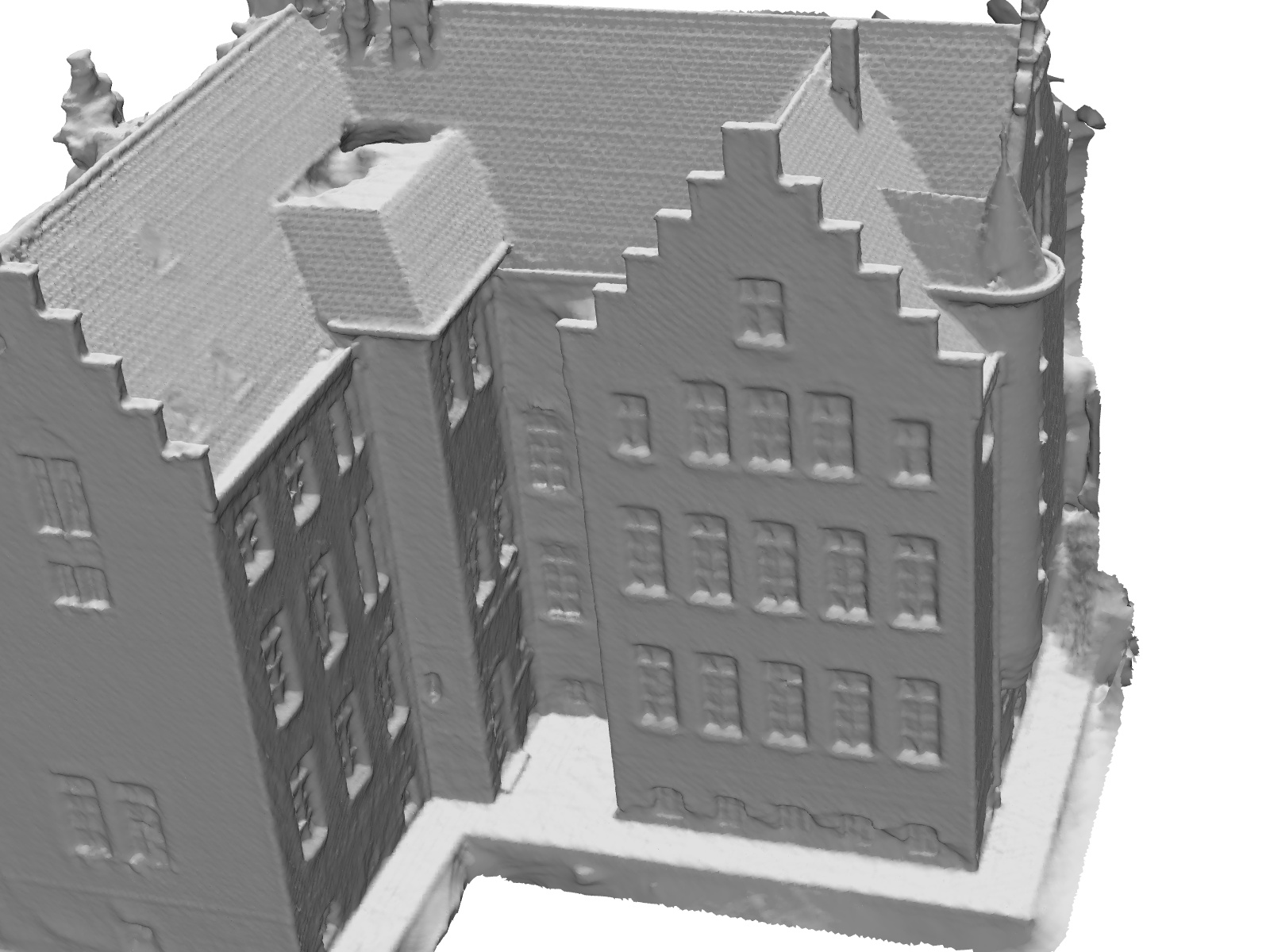}
  \caption*{Ours}}
   \caption{Ablation study on Scan 24 in DTU dataset. Notice the dent on the roof and left-most facade of the building.}
\label{fig:ablation}
\end{figure*}

To evaluate the effectiveness of our approach, we perform a qualitative ablation study on scene 24 of the DTU dataset (see Figure \ref{fig:ablation}). When we train our full model, the resulting surface retains overall geometric correctness and displays high amounts of surface detail. When we remove $\mathcal{S}_{coarse}$ surface regularization, the level of detail on the surface does not change, but the large-scale geometry becomes incorrect, resulting in the formation of a large bump protruding from the roof. When we remove coarse-to-fine annealing, surface detail quality is still similar to the full model, but a large crevice forms in the wall closest to the camera. Finally, when we directly train our model without surface regularization or coarse-to-fine annealing, the level of surface detail is still the same, but a lot more bumps and crevice appear in the large-scale geometry of the scene, making it much less accurate.

\section{Conclusion}
\label{sec:conclusion}
We propose \methodname, a novel neural implicit surface reconstruction algorithm that uses multi-resolution hash grid encoding, a special surface regularization technique, and coarse-to-fine training strategy. By incorporating these strategies, our method is able to capture significantly more high-frequency surface geometry compared to prior works. In terms of geometric accuracy, \methodname\ outperforms its predecessor work \cite{neus} on the DTU dataset \cite{dtu} and is on par with other works which utilize other input signals such as multi-view constraints \cite{francois2022neuralwarp}. One limitation to our work is that it is based on NeuS \cite{neus} and does not utilize multi-view constraints which other works \cite{francois2022neuralwarp, geoneus} utilize to improve overall geometric accuracy. However, these multi-view constraints are not at odds with the high-frequency details our algorithm recovers. Another limitation is that our algorithm does not address inherent ambiguity between the shading and surface normal on the input images. In the future, we would like to experiment on incorporating multi-view constraints and lighting/material estimation to further improve our work. 

\section*{Acknowledgements} 
We would like to thank Frederic Devernay, Karl Hillesland, Changgong Zhang, and Yu Lou for their insightful comments and suggestions. Erich Liang was funded in part by an NSF GRFP (grant \#2146752).

{\small
\bibliographystyle{ieee_fullname}
\bibliography{arXiv_copy}
}

\end{document}